\documentclass[11pt]{article}
\usepackage[final]{acl}

\usepackage{times}
\usepackage{latexsym}

\usepackage[T1]{fontenc}

\usepackage[utf8]{inputenc}

\usepackage{microtype}

\usepackage{inconsolata}

\usepackage{graphicx}

\usepackage[most]{tcolorbox}
\usepackage{array}
\usepackage{booktabs}
\usepackage{tabularx}
\usepackage{placeins}
\usepackage[dvipsnames,svgnames,x11names]{xcolor}
\usepackage{colortbl}
\usepackage{multirow,makecell}
\usepackage{cleveref}
\usepackage{epigraph}
\usepackage{pifont}
\usepackage{ragged2e}
\usepackage{arydshln}
\usepackage{tikz}
\usetikzlibrary{decorations.shapes}
\usepackage{amsmath}
\usepackage{amssymb}
\usepackage[normalem]{ulem}
\usepackage{hypcap}
\usepackage{capt-of}
\usepackage{fontawesome}
\usepackage{bm}
\usepackage{romannum}
\usepackage{atbegshi}
\usepackage{hyperref}
\usepackage{subfig}
\usepackage{titletoc}
\usepackage{marvosym}
\usepackage{xspace}

\definecolor{OliveGreen}{rgb}{0.0, 0.5, 0.0}
\definecolor{OliveGreen}{rgb}{0.0, 0.6, 0.0}
\definecolor{rred}{rgb}{0.75, 0.0, 0.0}
\definecolor{bblue}{rgb}{0.13, 0.67, 0.8}
\definecolor{sblue}{RGB}{240,248,255} 
\definecolor{sred}{RGB}{255,240,245} 
\definecolor{sgreen}{RGB}{239, 255, 232} 

\newcommand\coauth{$^\star$}
\newcommand{\correspond}{$^\dag$}

\newcolumntype{C}{>{\arraybackslash}X}
\newcommand{\redcross}{\textcolor{red}{\ding{56}}}
\newcommand{\greencheck}{\textcolor{green!60!black}{\ding{52}}}

\newcommand{\labeledbox}[2]{%
    \begin{tikzpicture}
        \node[draw, line width=0.6pt, rounded corners=4pt, inner sep=6pt, text width=0.85\linewidth] 
            (box) {#2};
        \node[rotate=90, anchor=center] 
            at ([xshift=-0.38cm]box.west) {\textbf{#1}};
    \end{tikzpicture}%
}
\newcommand{\taskcomp}{%
    \par%
    \makebox[\linewidth][c]{\hspace{2.6em}$\Updownarrow$ \quad \textit{Task Composition} \quad $\Updownarrow$}%
    \par\vspace{0.2em}%
}
\newcommand{\ourbenchmark}{\textsc{SemanticQA}\xspace}
\newcommand{\hlcell}{\cellcolor{AliceBlue!80}}
\newcommand{\hlcelll}{\cellcolor{LavenderBlush!80}}
\newcommand{\hlcellll}{\cellcolor{LightGreen!15}}

\newcommand{\blfootnote}[1]{%
  \begingroup
  \renewcommand\thefootnote{}\footnote{#1}%
  \addtocounter{footnote}{-1}%
  \endgroup
}

\newcommand{\hlinterpret}[1]{%
    \tikz[baseline]{\node[fill=AliceBlue!100, rounded corners=3.2pt, inner sep=2.4pt]{#1};}%
}
\newcommand{\hlclassify}[1]{%
    \tikz[baseline]{\node[fill=LavenderBlush!100, rounded corners=3.2pt, inner sep=3pt]{#1};}%
}
\newcommand{\hlextract}[1]{%
    \tikz[baseline]{\node[fill=LightGreen!25, rounded corners=3.2pt, inner sep=3pt]{#1};}%
}

\tikzset{
    decoration = {shape backgrounds, shape=circle, shape size=0.8pt, shape sep={2pt, between centers}},
    paint/.style = {decorate, fill=black}
}
\newtcolorbox{DottedTextBox}{
    enhanced,
    boxsep=0mm, left=4pt,
    arc=4pt,
    colback=white,
    colframe=white,
    pad at break=0pt,bottomrule at break=0pt,toprule at break=0pt,
    borderline={0pt}{0pt}{paint},
}
\title{Revisiting\textit{ a Pain in the Neck}:\\A Semantic Reasoning Benchmark for Language Models}

\author{
 Yang Liu$^{1,2}$\coauth{},
 Hongming Li$^{1}$\coauth{},
 \\
 \textbf{Melissa Xiaohui Qin$^{1}$, Qiankun Liu$^{1}$,
 Chao Huang$^{1}$\correspond{}}\vspace{0.1em}
\\
 \textsuperscript{1}University of Science and Technology Beijing,\vspace{0.04em}\\
 \textsuperscript{2}State Key Laboratory of General Artificial Intelligence, BIGAI\vspace{0.04em}\\
  liuyang@bigai.ai, hongmingli.lhm@gmail.com, \vspace{0.04em}\\ \{qinxiaohui, liuqk3, chaohuang\}@ustb.edu.cn
}

\begin{document}
\pagenumbering{arabic}
\setcounter{page}{1}
\maketitle

\begin{abstract}
We present \ourbenchmark, an evaluation suite designed to assess language models (LMs) in semantic phrase processing tasks. The benchmark consolidates existing multiword expression (\textsc{MwE}) resources and reorganizes them into a unified testbed. It covers both general lexical phenomena, such as lexical collocations, and three fine-grained categories: idiomatic expressions, noun compounds, and verbal constructions. Through \ourbenchmark, we assess LMs of diverse architectures and scales in extraction, classification, and interpretation tasks, as well as sequential task compositions. We reveal substantial performance variation, particularly on tasks requiring semantic reasoning, highlighting differences in reasoning efficacy and semantic understanding of LMs, providing insights for pushing LMs with stronger comprehension on non-trivial semantic phrases. The evaluation harness and data of \ourbenchmark are available at \faicon{github} \url{https://github.com/jacklanda/SemanticQA}.
\blfootnote{\coauth Equal contribution.}\blfootnote{\correspond Correspondence to: chaohuang@ustb.edu.cn}
\end{abstract}

\section{Introduction}\label{sec:introduction}
Semantic phrases (SP), also referred to as multiword expressions (\textsc{MwE}), are lexical combinations whose meanings or usages may not be fully derived from their individual components \citep{pasquer-etal-2020-verbal}. They exhibit varying degrees of compositionality, idiomaticity, and fixedness \citep{sailer2018multiword, ramisch2023multiword}. Despite extensive work in supervised and unsupervised paradigms, robust SP processing remains a fundamental challenge in NLP \citep{sag2002multiword,constant2017multiword,shwartz-dagan-2019-still,ramisch2023survey,tanner-hoffman-2023-mwe}.

\begin{figure}[ht]
    \centering
    \hspace*{-0.28cm}%
    \footnotesize
    \begin{minipage}{\linewidth}
        \centering
        \labeledbox{\hlinterpret{Interpretation}}{%
            \textbf{[Context]} It was not \textit{rocket science} in this case to
            determine that Goebbels was being cited to attack Trump, not to praise Nazis.
            \par\smallskip
            \textbf{[Idiom]} rocket science
            \par\smallskip
            \textbf{[Interpretation]} something complex and difficult
        }

        \taskcomp

        \labeledbox{\hlextract{Extraction}}{%
            \textbf{[Context]} It was not \textit{rocket science} in this case to determine that
            Goebbels was being cited to attack Trump, not to praise Nazis.
            \par\smallskip
            \textbf{[Extracted Idiom]} rocket science
        }

        \taskcomp
        \labeledbox{\hlclassify{Classification}}{%
            \textbf{[Context]} It was not \textit{rocket science} in this case to
            determine that Goebbels was being cited to attack Trump, not to praise Nazis.
            \par\smallskip
            \textbf{[Idiom]} rocket science
            \par\smallskip
            \textbf{[Choices]}\par
            (A) Projectile knowledge \hfill \redcross \par
            \textbf{(B) Difficult problem} \hfill \greencheck \par
            (C) Proper Noun \hfill \redcross \par
            (D) Meta Usage \hfill \redcross
        }

    \end{minipage}

    \caption{Atomic task exemplars of idiomatic expression in \ourbenchmark, grouped as task compositions.}
    \label{fig:examples}
\end{figure}

\begin{figure*}[t]
  \centering
  \includegraphics[width=\linewidth]{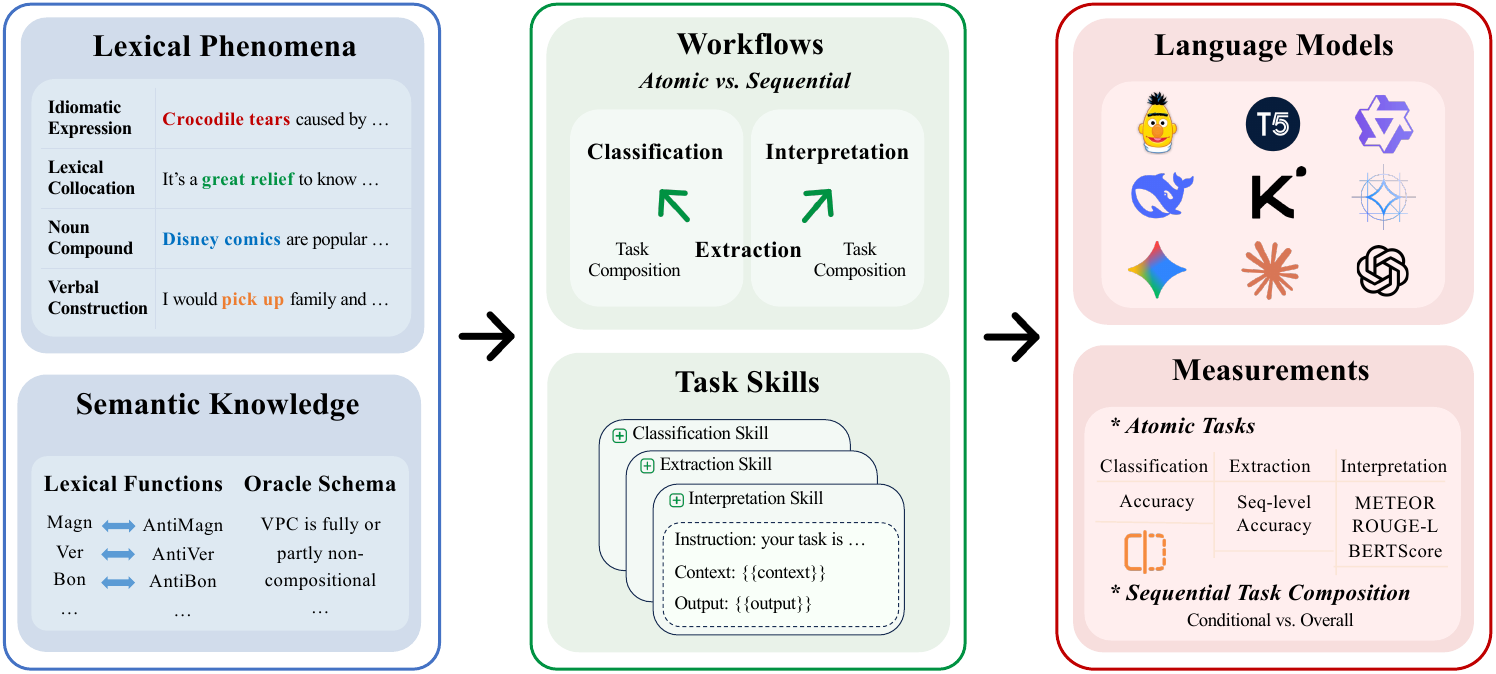}
  \caption{Overview of \ourbenchmark for benchmarking LMs on lexical phenomena.}
  \label{fig:experiments}
\end{figure*}

Language models (LMs) are typically evaluated using benchmarks that emphasize mathematical reasoning \citep{an2025amobenchlargelanguagemodels,balunovic_srimatharena_2025}, code generation \citep{Austin2021ProgramSW, NEURIPS2024_6a059625}, and logical reasoning \citep{li2025reflectevo,liu2026rulereasoner}. While these benchmarks effectively assess reasoning capacity and factual knowledge proficiency of LMs \citep{ICLR2024_7b7d7985,li2025reflectevo,yang2026onemillion}, they largely overlook fine-grained semantic reasoning that operates over sub-sentential units. In particular, phrasal semantics, where meaning emerges from interactions between lexical constituents and context, remain under-explored and, when evaluated, are often assessed via isolated task formats that conflate multiple semantic operations \cite{liu-etal-2026-lm}. As a result, it is difficult to determine whether strong performance reflects stable phrase-level semantic representations or task-specific heuristics. Therefore, recent work has called for diagnostic evaluations that disentangle semantic operations and investigate phrasal semantic behaviours beyond the understanding of superfacial language \citep{miletic2024semantics}.




We therefore ask: \textbf{How do language models behave when evaluated on phrasal semantics across distinct but structurally constrained task operations?} To answer this question, we introduce \ourbenchmark, an operation-aligned benchmark for semantic phrase processing. We adopt a deliberately operationalized view of semantic reasoning with respect to evaluation. Rather than requiring LMs to perform operations on the same instance, we examine whether phrasal semantic understanding generalizes across tasks that instantiate different operations. Specifically, we consider three atomic operations, including classification, extraction, and interpretation, which target the same underlying notion of phrasal sense while imposing structural constraints on LM generations. Under this formulation, semantic reasoning is assessed by the ability of LMs to exhibit compatible behavioral patterns across tasks, reflecting whether learnt phrasal semantics transfer across operations rather than overfitting to isolated task formats.


Under this definition, performance on a single task is insufficient. Instead, semantic reasoning is assessed via cross-operation consistency on \ourbenchmark, sensitivity to structural constraints, and robustness under compositional setups where applicable.
Our contributions are threefold:
\begin{enumerate}
  \item \textbf{Operation-aligned Semantic Evaluation.} \ourbenchmark does not introduce new semantic theories but evaluates phrasal competence through a set of controlled semantic operations with varying structural constraints. Its core contribution lies in \textbf{aligning existing SP tasks with the semantic operations they instantiate}, enabling systematic analyses of semantic behaviors across structural distinction yet related task families.

  \item \textbf{Minimal and Controlled Design.} \ourbenchmark employs fixed prompt templates to reduce prompt-induced variance across LMs. By holding prompt structure constant while varying semantic operations, it supports fair comparison under shared conditions.

  \item \textbf{Diagnostic Analyses of Cascade Sensitivity.} In explicitly designed sequential task setups, we show that strong LMs often fail to maintain semantic consistency across dependent operations, revealing phrase-level limitations that remain hidden in single-task evaluations.
\end{enumerate}

\section{Related Work}
\paragraph{Complex Reasoning.}
Recent work evaluated LMs in a wide range of areas \citep{NEURIPS2024_6a059625,an2025amobenchlargelanguagemodels,balunovic_srimatharena_2025}. They focus on structured reasoning over explicit representations, such as compositional procedures in math or symbolic tasks. Although effective for formal reasoning, they overlook fine-grained semantic operations and omit their applications in context \citep{liu2024mathbenchevaluatingtheoryapplication, an2025amobenchlargelanguagemodels, luong-etal-2025-towards}. 

In contrast, semantic reasoning relies on the composition of phrasal meaning, contextual disambiguation, semantic-role inference, and paraphrase mapping. These aspects require the manipulation of latent semantic representations rather than symbolic rules. Prior work shows that even frontier LMs often depend on shallow heuristics, implying the need for assessment to examine semantic reasoning \citep{10.1145/3664194, huang-etal-2025-structfact}.

\begin{figure}[t]
\centering
  \includegraphics[width=\columnwidth]{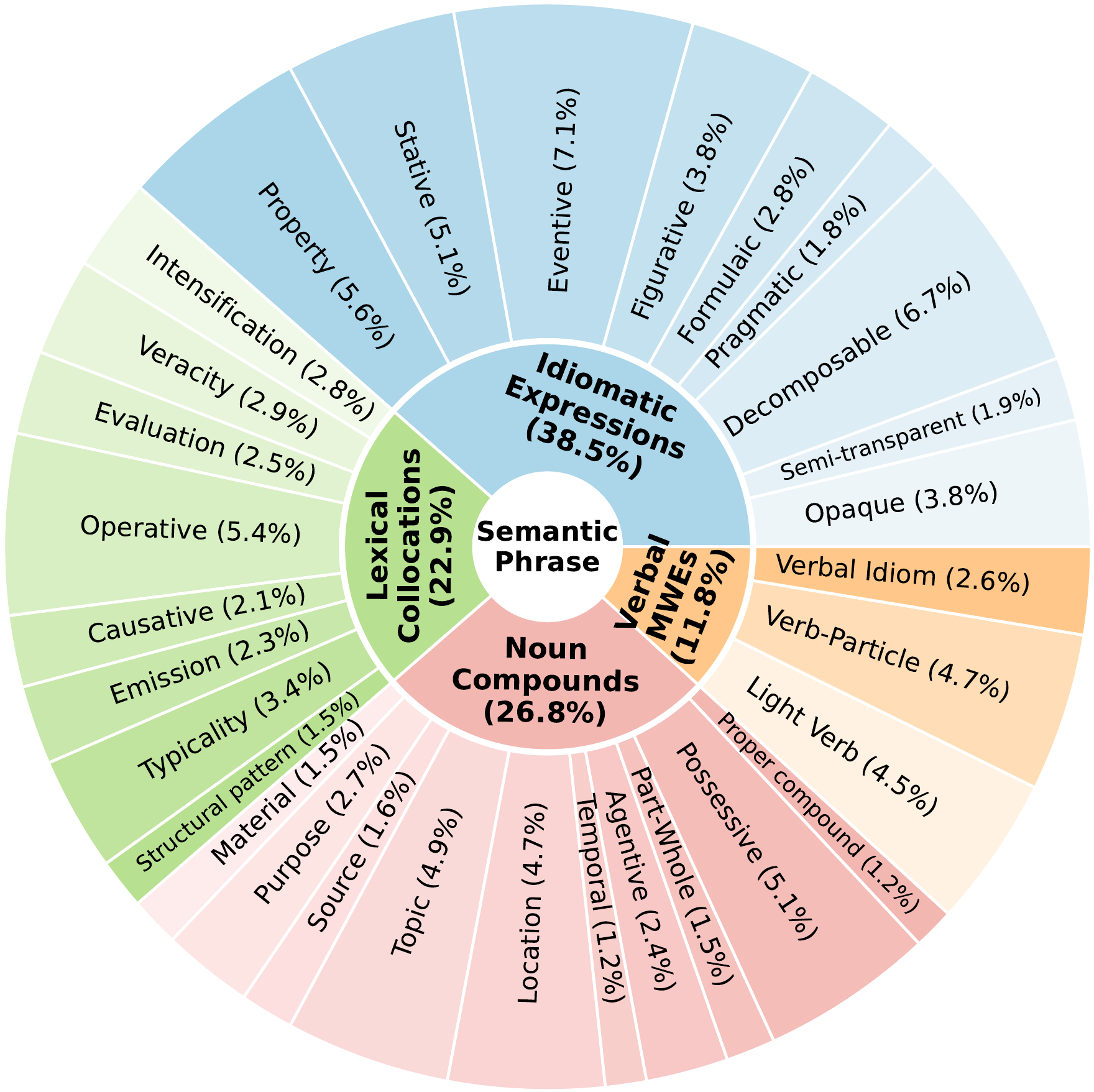}
\caption{The coverage of coarse- and fine-grain semantic phrase categories of \ourbenchmark. The mapping of semantic phrase categories is listed in Table \ref{tab:sp_mwe_mapping}.}
  \label{fig:semantic_phrase_taxonomy}
\end{figure}

\paragraph{Semantic Phrase Processing and Resources.}
Semantic phrase processing has long been studied, with early work focusing on unsupervised phrase representations and compositional modeling \citep{vacareanu-etal-2020-unsupervised, arase-tsujii-2020-compositional}. 
Recent work examined idiom identification, contextual paraphrasing, and noun compound interpretation using  transformers \citep{klubicka-etal-2023-idioms, wada-etal-2023-unsupervised}. 
In parallel, a wide range of data resources was developed to evaluate phrasal semantics, covering idiomatic expressions \citep{tedeschi-etal-2022-id10m,zhou-etal-2021-pie}, lexical collocations \citep{espinosa-anke-etal-2019-collocation,fisas-etal-2020-collfren,espinosa-anke-etal-2021-evaluating}, and verbal constructions \citep{savary-etal-2023-parseme,ramisch-etal-2020-edition}. 

Existing work typically isolates specific phrase types, task formats, and semantic phenomena, such as compositionality and idiomatic distinctions, without explicitly modeling the atomic semantic operations underlying phrase comprehension \citep{pham-etal-2023-pic,buijtelaar-pezzelle-2023-psycholinguistic,zeng-etal-2023-iekg}. Consequently, evaluations are conducted in isolation, limiting cross-task and cross-phenomenon analyses and generalization. This fragmentation motivates \ourbenchmark: a unified and operation-aligned benchmark for evaluating phrase-level semantic processing in LMs.

\begin{table*}
\vspace{-0.5em}
\centering
\tiny
\setlength\extrarowheight{2pt}
\begin{tabularx}{0.94\linewidth}{p{1.8cm}p{2.1cm}>{\centering\arraybackslash}p{1.2cm}>{\centering\arraybackslash}p{2.2cm}p{1.1cm}>{\centering\arraybackslash}p{1.9cm}>{\centering\arraybackslash}c}
\toprule
\textbf{Task} & \textbf{Data Source} & \textbf{Input $(\mathcal{I})$} &
\textbf{Output $(\mathcal{O})$} & \textbf{Metrics} &
\textbf{\# Test Size} &
\textbf{Phrase Type} \\
\midrule

IE Detection & \citet{tayyar-madabushi-etal-2021-astitchinlanguagemodels-dataset} &
$\mathcal{P} \oplus \mathcal{S} \oplus \mathcal{IE}$ & Choice from \textit{Options} &
\textsc{Acc} & 273 & \textsc{Idiomacity} \\
\midrule

IE Extraction & \citet{tedeschi-etal-2022-id10m} &
$\mathcal{P} \oplus \mathcal{S}$ & Extracted $\mathcal{IE}$ &
$\textsc{Acc}_s$ & 447 & \textsc{Idiomacity} \\
\midrule

IE Interpretation & \citet{zhou-etal-2021-pie, chakrabarty2022s} &
$\mathcal{P} \oplus \mathcal{S} \oplus \mathcal{IE}$ & Interpretation of $\mathcal{IE}$ &
\makecell[l]{\textsc{METEOR},\textsc{Rouge-L},\\\textsc{BERTScore}} &
818 & \textsc{Idiomacity} \\
\midrule

LC Categorization & \citet{espinosa-anke-etal-2021-evaluating} &
$\mathcal{P} \oplus \mathcal{T} \oplus \mathcal{S}$ & Choice from \textit{Options} &
$\textsc{Acc}$ &
305 & \textsc{Collocation} \\
\midrule

LC Extraction & \citet{fisas-etal-2020-collfren} &
$\mathcal{P} \oplus \mathcal{T} \oplus \mathcal{S}$ &
Extracted $\mathcal{LC}$ & $\textsc{Acc}_s$ &
305 & \textsc{Collocation} \\
\midrule

LC Interpretation &
\citet{espinosa-anke-etal-2019-collocation,espinosa-anke-etal-2021-evaluating} &
$\mathcal{P} \oplus \mathcal{S} \oplus \mathcal{LC}$ &
Interpretation of $\mathcal{LC}$ &
\makecell[l]{\textsc{METEOR},\textsc{Rouge-L},\\\textsc{BERTScore}} &
305 & \textsc{Collocation} \\
\midrule

NC Compositionality & \citet{garcia-etal-2021-assessing} &
$\mathcal{P} \oplus \mathcal{S} \oplus \mathcal{NC}$ &
Choice from \textit{Options} & $\textsc{Acc}$ &
242 & \textsc{Noun Compound} \\
\midrule

NC Extraction & \citet{garcia-etal-2021-assessing,kolluru-etal-2022-covid} &
$\mathcal{P} \oplus \mathcal{S}$ & Extracted $\mathcal{NC}$ & $\textsc{Acc}_s$ &
720 & \textsc{Noun Compound} \\
\midrule

NC Interpretation & \citet{coil-shwartz-2023-chocolate} &
$\mathcal{P} \oplus \mathcal{S} \oplus \mathcal{NC}$ &
Interpretation of $\mathcal{NC}$ &
\makecell[l]{\textsc{METEOR},\textsc{Rouge-L},\\\textsc{BERTScore}} &
110 & \textsc{Noun Compound} \\
\midrule

\textsc{VMwE} Extraction & \citet{savary-etal-2023-parseme} &
$\mathcal{P} \oplus \mathcal{S}$ & Extracted $\mathcal{VC}$ &
$\textsc{Acc}_s$ & 475 & \textsc{Verbal MwE} \\

\bottomrule
\end{tabularx}
\caption{A summary of the data statistics in \ourbenchmark. $\mathcal{P}$ refers to the prompt template, $\mathcal{S}$ denotes the sentence context, $\mathcal{T}$ represents the semantic taxonomy narrative, $\mathcal{IE}$ denotes idiomatic expressions, $\mathcal{LC}$ denotes lexical collocations, and $\mathcal{NC}$ denotes noun compounds.}
\label{tab:task-list}
\end{table*}

\section{\ourbenchmark}\label{sec:lexbench} 
\subsection{Preliminaries}
Semantic phrase exhibit diverse degrees of compositionality and idiomacity. We consider four representative phrase types that capture popular sources of phrase variation. In addition, drawing on prior literature, we systematically consolidate and standardize the scope of SP considered in this work. We also employ LMs to label fine-grained categories of the phrases, with the resulting classification illustrated in Figure~\ref{fig:semantic_phrase_taxonomy} \citep{54311d36-f6a7-378f-b3bb-f2b37bdfb9f1,10.1007/3-540-45715-1_1,tratz-hovy-2010-taxonomy,savary-etal-2017-parseme,tayyar-madabushi-etal-2021-astitchinlanguagemodels-dataset,kolluru-etal-2022-covid,chakrabarty-etal-2022-rocket,mel2023general}.

\paragraph{Lexical Collocations (LC).} LC forms a broad class of SPs with varying degrees of compositionality. They are characterized by conventionalized lexical relations between a \textit{base} word and a \textit{collocate} word, ranging from largely compositional combinations to idiom-style usages \citep{espinosa-anke-etal-2021-evaluating,math10203831}.

\paragraph{Idiomatic Expressions (IE).}
IE are prototypical non-compositional phrases whose meanings can not be derived from their constituent words (\textit{e.g.}, \textit{kick the bucket}). Processing such expressions requires LMs to recover conventionalized meanings beyond literal composition. \citep{zhou2022idiomatic,zeng-bhat-2022-getting,haviv-etal-2023-understanding}.

\paragraph{Noun Compounds (NC).}
NC are often compositional, but their interpretation frequently depends on implicit semantic relations, contextual cues, or world knowledge (\textit{e.g.}, \textit{baby oil} \textit{vs.} \textit{olive oil}) \citep{kolluru-etal-2022-covid,coil-shwartz-2023-chocolate}.

\paragraph{Verbal Constructions (VC).}
VC or verbal multiword expressions (\textsc{VMwE}), including light-verb constructions (LVC), verb–particle constructions (VPC), and verbal idioms (VID), are typically semi-compositional \citep{tanner-hoffman-2023-mwe,savary-etal-2023-parseme,ramisch-etal-2023-survey}. Their meanings arise from an interaction between literal composition and conventional usage.


\subsection{Benchmark Construction} 
\ourbenchmark\ is built upon prior resources \citep{tayyar-madabushi-etal-2021-astitchinlanguagemodels-dataset, espinosa-anke-etal-2022-multilingual, garcia-etal-2021-assessing, savary-etal-2023-parseme}, which vary in annotation protocols, difficulty distributions, and semantic granularity. Rather than enforcing uniform difficulty or annotation consistency across sources, \ourbenchmark is designed to reflect the variation and is not intended for absolute comparisons on phrase types. We focus on within-task trends, as well as relative changes induced by semantic operations and sequential compositions.
Semantic reasoning is grounded in performance patterns that are stable across multiple tasks and datasets, rather than in absolute scores. All experiments use the datasets described in Table~\ref{tab:task-list} and \S\ref{sec:appendix-additional-experiment-details}.


\begin{figure*}[t]
    \centering
    \includegraphics[width=0.8\linewidth]{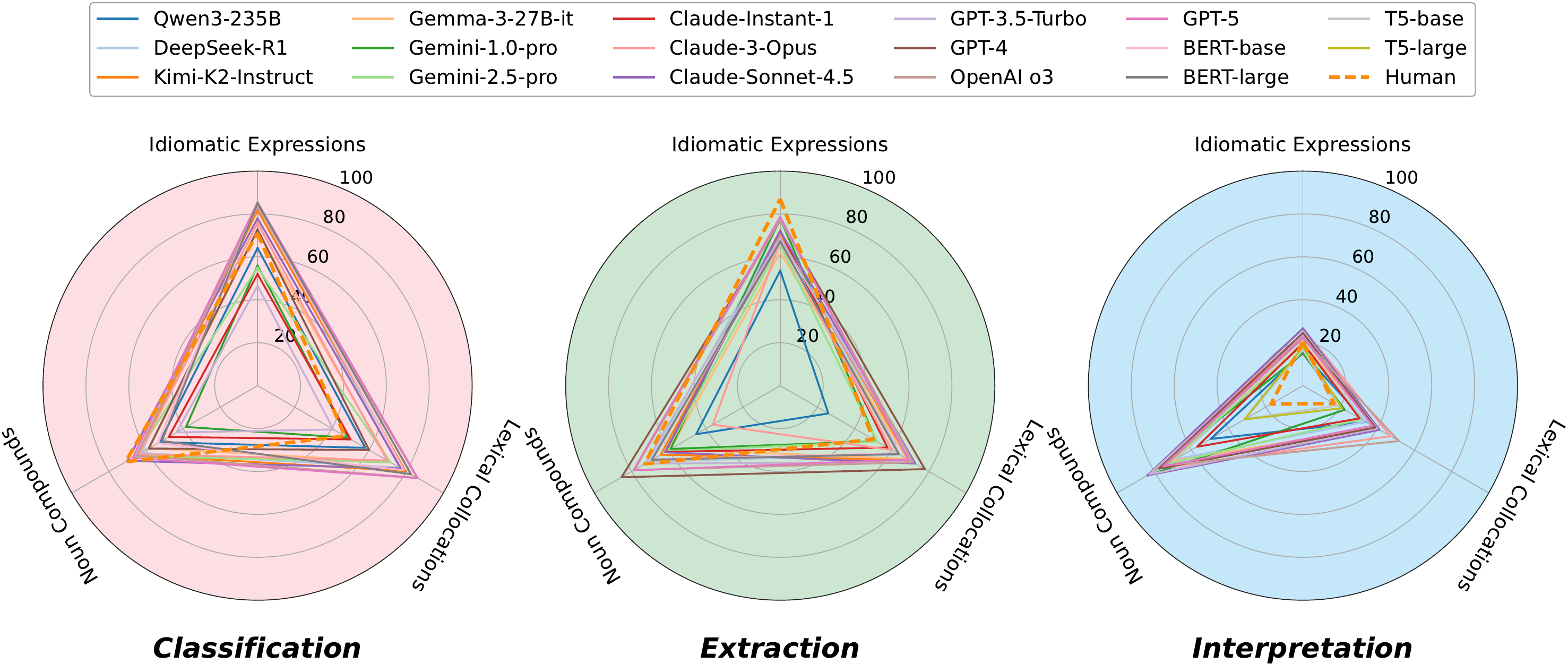}
    \caption{Overall the best performance (\textit{i.e.}, capacity triangle $\triangle$) of models on \ourbenchmark}
    \label{fig:Performance-SpiderFigure}
\end{figure*}

\begin{table}
\centering
\scriptsize
\begin{tabular}{l@{\hspace{-2.5em}}c@{\hspace{-1.4em}}r}
\toprule
\textbf{Lexical Function} & \textbf{Example} & \textbf{Semantic Relation} \\
\midrule
Magn & Magn(\textit{rain}) = \textit{heavy} & ``intense'', ``strong'' \\
AntiMagn & AntiMagn(\textit{accent}) = \textit{slight} & ``little'', ``weak'' \\
\midrule
Ver & Ver(\textit{message}) = \textit{clear} & ``real'', ``genuine''\\
AntiVer & AntiVer(\textit{accusation}) = \textit{groundless} & ``non-genuine'' \\
\midrule
Bon & Bon(\textit{bread}) = \textit{fresh} & ``positive'' \\
AntiBon & AntiBon(\textit{advantage}) = \textit{undue} & ``negative'' \\
\midrule
Son & Son({alarm clock}) = \textit{ring(s)} & ``sound'', ``voice'' \\
\midrule
Oper1 & Oper1(\textit{advice}) = \textit{give} & ``perform'' \\
\bottomrule
\end{tabular}
\caption{Partial semantic relations involved in this paper, with their exemplars. More relations in lexical functions (LFs) can be referred to the Table \ref{tab:lexfunc-with-semantic-gloss}.}
\label{tab:collocation-semantic-relations}
\end{table}

\subsection{Task Definitions} \label{subsec:Task-Definition}
We organize tasks by both phrase types and atomic task operations, where each operation targets a distinct aspect. This allows tasks operating on the same underlying phrase meaning to differ in their output structure and constraints.

For IE, we include detection (IED), extraction (IEE), and interpretation (IEI) tasks. Detection is formulated as a multiple-choice classification task, extraction requires exact span identification, and interpretation evaluates contextualized paraphrase generation. All datasets are adapted from existing annotated resources \citep{tayyar-madabushi-etal-2021-astitchinlanguagemodels-dataset,tedeschi-etal-2022-id10m, zhou-etal-2021-pie}, with overlapping instances deduplicated and reformatted to ensure consistency across operations.

For LC, we design categorization (LCC), extraction (LCE), and interpretation (LCI) tasks. Categorization requires predicting the semantic relation of a collocation under a lexical function taxonomy (\textit{cf.} Table~\ref{tab:collocation-semantic-relations} and Appendix~\S \ref{sec:appendix-lf_semantic_gloss})  \citep{mel2023general}. Extraction identifies both the \textit{base} word and \textit{collocate} word in context, while interpretation (\textit{cf.} Appendix~\S\ref{sec:annotation-guideline}) evaluates paraphrasing conditioned on context. Datasets are balanced across semantic relation categories to support controlled multi-class evaluation \citep{espinosa-anke-etal-2021-evaluating, fisas-etal-2020-collfren,espinosa-anke-etal-2022-multilingual, espinosa-anke-etal-2021-evaluating}.

For NC, we include compositionality classification (NCC), extraction (NCE), and interpretation (NCI) tasks, which evaluate compositionality judgement, structural identification, and literal meaning reconstruction in a given context, respectively \citep{garcia-etal-2021-assessing, kolluru-etal-2022-covid, coil-shwartz-2023-chocolate, hendrickx-etal-2013-semeval}.

For \textsc{VMwE}, we include \textsc{VMwE} extraction task, which requires identifying a single verbal construction in context, covering VPC (VPE), LVC (LVE), and VID (VIE) \citep{savary-etal-2023-parseme}.

Finally, we formalize SP processing as a conditional generation problem under operation constraints. Given a prompt template $\mathcal{P}$ (\textit{cf.} Appendix \S\ref{sec:example-prompt}) that specifies a target operation and an SP embedded in its context $\mathcal{S}$, a LM is required to generate an output $\mathcal{O}$ that satisfies the instruction induced by $\mathcal{P}$. Concretely, the model input is constructed as $\mathcal{I} := \mathcal{P} \oplus \mathcal{S}$, where $\oplus$ denotes a task-specific composition of instruction and contextualized phrase. The output $\mathcal{O}$ varies according to the semantic operation being evaluated. For example, in extraction tasks, $\mathcal{O}$ corresponds to the target phrase span identified from $\mathcal{S}$ under the constraints specified by $\mathcal{P}$, whereas in classification or interpretation tasks, $\mathcal{O}$ represents a semantic decision or reconstruction aligned with the given instruction.

For each task $t$, the configuration of the tuple $(\mathcal{P}, \mathcal{S}, \mathcal{O})$ is instantiated according to a fixed template, as in Table~\ref{tab:task-list}. The dataset for task $t$ is defined as $\mathcal{D}^{(t)} := \{(p^{(t)}, s_i^{(t)}, o_i^{(t)})\}_{i=1}^{N}$, where each example pairs a prompt, a contextualized SP, and a gold-standard output corresponding to the target semantic operation (\textit{cf.} Figures~\ref{fig:examples} and \ref{fig:experiments}).


\begin{table*}[!ht]
\centering
\scriptsize 
\setlength\tabcolsep{4pt}
\setlength{\extrarowheight}{2pt}
\begin{tabular}{lcccccccccccccccc}
\toprule
\multirowcell{2.7}{\textbf{{\scriptsize{M}}ODEL}} & \multicolumn{3}{c}{\textbf{{\scriptsize{I}}DIOM}} & \phantom{c} & \multicolumn{3}{c}{\textbf{{\scriptsize{C}}OLLOCATION}} & \phantom{c} & \multicolumn{3}{c}{\textbf{{\scriptsize{N}}OUN {\scriptsize{C}}OMPOUND}} & \phantom{c} & \multicolumn{3}{c}{\textbf{{\scriptsize{VM}}W{\scriptsize{E}}}} & \phantom{c} \\
\cmidrule{2-4}\cmidrule{6-8}\cmidrule{10-12}\cmidrule{14-16}
& \textbf{IED} & \textbf{IEE} & \textbf{IEI} && \textbf{LCC} & \textbf{LCE} & \textbf{LCI} && \textbf{NCC} & \textbf{NCE} & \textbf{NCI} && \textbf{VPE} & \textbf{LVE} & \textbf{VIE} & \\
\midrule
\textbf{{\textsc{Metric}}~\raisebox{0.23ex}{$(\%)$}} & $\textsc{Acc}$ & \hspace{0.05cm}$\textsc{Acc}_s$ & \textsc{Mtr} && \textsc{Acc} & \hspace{0.05cm}$\textsc{Acc}_s$ & \textsc{Mtr} && $\textsc{Acc}$ & \hspace{0.05cm}$\textsc{Acc}_s$ & \textsc{Mtr} && \hspace{0.05cm}$\textsc{Acc}_s$ & \hspace{0.05cm}$\textsc{Acc}_s$ & \hspace{0.05cm}$\textsc{Acc}_s$ \\
\midrule
\hlcelll \textsc{Human} & 71.0 & \uline{\textbf{87.0}} & 20.5 && 47.0 & 50.0 & 16.7 && \uline{\textbf{71.0}} & 73.0 & 17.2 && \uline{\textbf{85.0}} & \uline{\textbf{55.0}} & \uline{\textbf{78.0}} \\
\midrule

\hlcellll \textsc{DeepSeek-R1}: \textit{zero-shot} & 71.1 & 69.4 & \textcolor{rred}{\textbf{12.4}} && 66.6 & \textcolor{rred}{\textbf{31.5}} & \textcolor{rred}{\textbf{31.8}} & & 60.2 & 51.3 & \textcolor{rred}{\textbf{31.4}} && 76.8 & 26.7 & 50.5 & \\
\hlcellll \hspace{0.88cm}$\hookrightarrow \text{+}$ \textit{three-shot} & 79.1 & 70.6 & 19.4 & & 76.4 & 55.6 & 33.6 & & 62.7 & 66.3 & 68.3 & & 74.7 & 26.7 & \textcolor{OliveGreen}{\textbf{59.1}} & \\
\hlcellll \hspace{0.88cm}$\hookrightarrow \text{+}$ \textit{five-shot} & 84.3 & 72.3 & 19.2 & & 76.1 & 64.3 & 32.9 & & 60.6 & 70.7 & 68.7 & & 81.6 & 35.8 & 57.1 & \\
\hlcellll \textsc{Kimi-K2-Instruct}: \textit{zero-shot} & 68.5 & 63.1 & 13.9 & & 68.5 & 34.4 & 33.7 & & 60.6 & 45.4 & 65.4 & & \textcolor{rred}{\textbf{55.8}} & 28.9 & 46.7 & \\
\hlcellll \hspace{0.59cm}$\hookrightarrow \text{+}$ \textit{three-shot} & 77.7 & 68.9 & 23.5 & & 79.0 & 67.9 & 39.1 & & 59.3 & 64.4 & 71.4 & & 79.5 & 39.4 & 43.8 & \\
\hlcellll \hspace{0.59cm}$\hookrightarrow \text{+}$ \textit{five-shot} & 81.7 & 69.6 & 21.7 & & 79.7 & 69.2 & 36.9 & & 64.7 & 63.6 & 76.7 & & 81.1 & \textcolor{OliveGreen}{\textbf{43.3}} & 46.7 & \\
\hlcellll \textsc{Gemma-3-27B-it}: \textit{zero-shot} & \textcolor{rred}{\textbf{55.0}} & \textcolor{rred}{\textbf{57.3}} & 13.5 & & \textcolor{rred}{\textbf{58.0}} & 38.4 & 35.0 & & 58.3 & \textcolor{rred}{\textbf{39.9}} & 43.8 & & 66.8 & 19.4 & 38.1 & \\
\hlcellll \hspace{0.59cm}$\hookrightarrow \text{+}$ \textit{three-shot} & 69.6 & 62.0 & 19.9 & & 70.1 & 63.7 & 37.3 & & 56.7 & 57.2 & 68.3 & & 74.1 & 28.3 & 45.7 & \\
\hlcellll \hspace{0.59cm}$\hookrightarrow \text{+}$ \textit{five-shot} & 72.1 & 61.6 & 19.2 & & 70.8 & 68.2 & 38.7 & & 56.2 & 59.2 & 70.5 & & 70.5 & 35.0 & 52.4 & \\

\hdashline
 \hlcell \textsc{Claude-Sonnet-4.5}: \textit{zero-shot} & 72.5 & 68.5 & 17.0 & & 67.5 & 40.1 & 34.8 & & \textcolor{rred}{\textbf{51.0}} & 45.1 & 77.2 & & 69.8 & \textcolor{rred}{\textbf{16.1}} & \textcolor{rred}{\textbf{41.9}} & \\
\hlcell \hspace{1cm}$\hookrightarrow \text{+}$ \textit{three-shot} & 77.7 & 72.0 & 25.8 & & 77.1 & 70.5 & 41.2 & & 61.4 & 59.3 & 81.2 & & 76.8 & 30.6 & 42.9 & \\
\hlcell \hspace{1cm}$\hookrightarrow \text{+}$ \textit{five-shot} & 78.0 & 72.0 & \textcolor{OliveGreen}{\textbf{26.7}} & & 76.1 & \textcolor{OliveGreen}{\textbf{72.7}} & 40.8 & & \textcolor{OliveGreen}{\textbf{70.1}} & 62.1 & \textcolor{OliveGreen}{\textbf{83.8}} & & \textcolor{OliveGreen}{\textbf{82.0}} & 37.2 & 47.6 & \\

\hlcell \textsc{OpenAI o3}: \textit{zero-shot} & 57.1 & 65.1 & 12.6 && 72.1 & 37.7 & 35.9 & & 65.2 & 62.9 & 45.7 && 67.9 & 25.6 & 51.4 & \\
\hlcell \hspace{0.88cm}$\hookrightarrow \text{+}$ \textit{three-shot} & 79.5 & 77.4 & 21.3 & & 85.9 & 65.3 & \textcolor{OliveGreen}{\textbf{41.6}} & & 58.9 & 77.5 & 68.2 & & 76.3 & 29.1 & 52.4& \\
\hlcell \hspace{0.88cm}$\hookrightarrow \text{+}$ \textit{five-shot} & 83.5 & 74.7 & 21.9 & & 83.6 & 71.5 & 35.9 & & 63.5 & 78.6 & 74.5 & & 77.3 & 36.9 & 50.0 & \\
\hlcell \textsc{GPT-5}: \textit{zero-shot} & 82.8 & 67.6 & 13.9 & & 75.4 & 36.7 & 33.7 & & 66.8 & 64.3 & 57.3  & & 74.2 & 28.9 & 56.2 & \\
\hlcell \hspace{0.88cm}$\hookrightarrow \text{+}$ \textit{three-shot} & 82.1 & 78.3 & 22.6 & & \textcolor{OliveGreen}{\textbf{86.2}} & 67.2 & 35.4 & & 61.8 & 77.1 & 70.1 & & 74.7 & 33.3 & 51.4 & \\
\hlcell \hspace{0.88cm}$\hookrightarrow \text{+}$ \textit{five-shot} & \textcolor{OliveGreen}{\textbf{85.4}} & \textcolor{OliveGreen}{\textbf{78.7}} & 22.5 & & 84.3 & 68.9 & 37.4 & & 67.2 & \textcolor{OliveGreen}{\textbf{79.0}} & 75.3 & & 74.7 & 38.3 & 50.5 & \\

\bottomrule
\end{tabular}
\caption{Major experimental results on \ourbenchmark. 
\uline{\textbf{Digits}} highlight cases in which human scores are higher than those of all evaluated models, serving as a coarse reference. \colorbox{sred}{Light Pink} indicates the human baseline. \colorbox{sgreen}{Light Green} and \colorbox{sblue}{Light Blue} present open-source models and proprietary models, respectively.
Green indicates the highest performance across all models (zero-shot and few-shot) within each task category, while red indicates the lowest.
}
\label{tab:major-exp-results}
\end{table*}

\subsection{Measurement}
We adopt task-appropriate automatic metrics aligned with the output characteristics of each semantic operation. Classification tasks are evaluated using accuracy (\textsc{Acc}). Extraction tasks are evaluated using the accuracy of the exact match at the sequence-level ($\textsc{Acc}_s$), which requires the exact recovery of the target phrase from the given context and avoids the inflation of the score from partial matches. Interpretation tasks are evaluated using \textsc{Meteor} (MTR) \citep{denkowski-lavie-2014-meteor} as the primary metric, with ROUGE-L (R-L) \citep{lin-2004-rouge} and BERTScore (B-S) \citep{zhang2019bertscore} reported for complementary analyses.

\section{Results}

\subsection{Evaluation Setups}
We evaluate a diverse set of LMs spanning different architectures, scales, and reasoning capabilities (\textit{cf.} Appendix~\S\ref{sec:additional-exp-details} and Tables~\ref{tab:models} and \ref{tab:hyperparams-bert-t5}), including GPT-5 \citep{openai_gpt5}, OpenAI o3 \citep{openai_o3}, GPT-4 \citep{openai2023gpt4}, Claude-Sonnet-4.5 \citep{Anthropic_Claude}, Gemini-2.5-Pro \citep{google_gemini}, Claude-3-Opus \citep{TheC3}, DeepSeek-R1 \citep{guo2025deepseek}, Qwen3-235B \citep{qwen3technicalreport}, Gemma-3-27B-it \citep{gemma_2025}, and Kimi-K2-Instruct \citep{kimiteam2025kimik2openagentic}, BERT-base/large \citep{devlin2019bert}, and T5-base/large \citep{raffel2020exploring}, as summarized in Figure~\ref{fig:Performance-SpiderFigure} and Table~\ref{tab:major-exp-results—appendix}. 

\subsection{Benchmarking Results}
\begin{figure*}[t]
\centering
  \includegraphics[width=\linewidth]{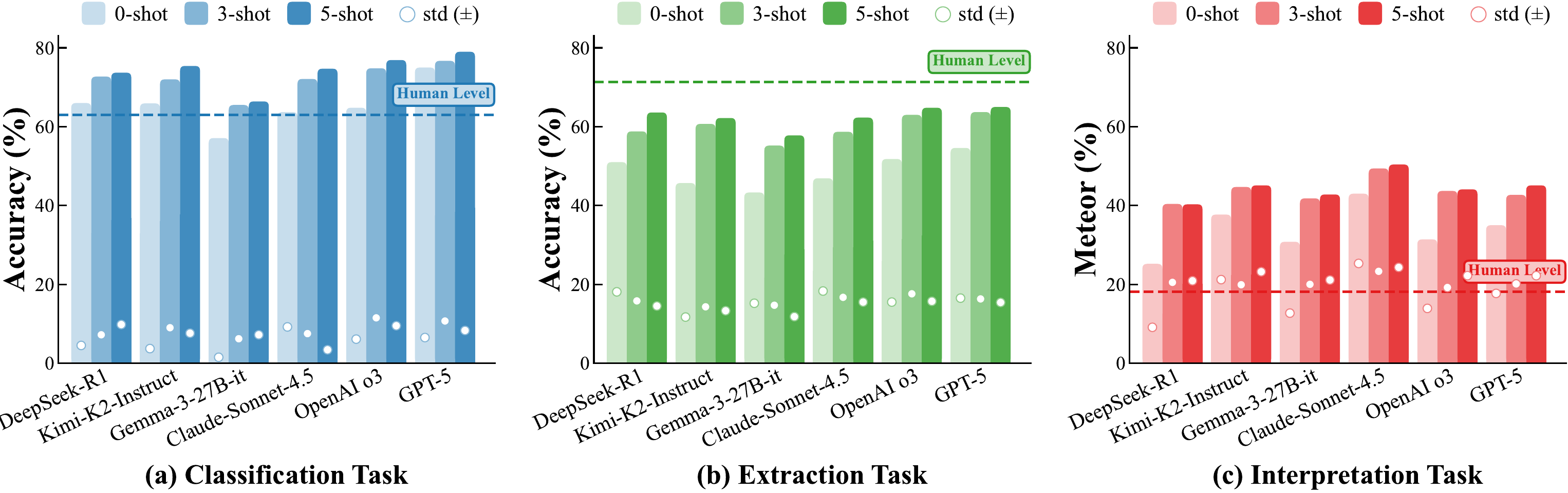}
  \caption{Grouped bars represent the mean performance of each model, while circular markers denote the population standard deviation (std), computed across tasks within the same category.
  }
  \label{fig:avg-std}
\end{figure*}

\paragraph{Overall Performance Patterns.}
Table~\ref{tab:major-exp-results}, Figures~\ref{fig:avg-std} and \ref{fig:Performance-SpiderFigure} show substantial variation across operations and phrase types (see Tables~\ref{tab:major-exp-results—appendix} and \ref{tab:vmwe-full-results}). Even within the same lexical phenomenon (\textit{e.g.}, IE or LC), models behave differently in tasks, indicating that these operations impose distinct structural and semantic constraints. Interestingly, no model performs uniformly well across all setups, suggesting operation-specific strengths and weaknesses rather than a single transferable notion of phrase-level competence. Moreover, \ourbenchmark is neither saturated nor uniformly difficult: different tasks expose complementary failure modes, supporting its use as a diagnostic testbed rather than a leaderboard driven by aggregating scores.

\begin{table}[t]
\renewcommand{\arraystretch}{0.9}
\scriptsize
\centering
\begin{tabularx}{\linewidth}{p{2.0cm} *{6}{>{\centering\arraybackslash}X}}
\toprule
\multirow{2.7}{*}{\textbf{Model}} 
& \multicolumn{2}{c}{\textbf{IEI}} 
& \multicolumn{2}{c}{\textbf{LCI}} 
& \multicolumn{2}{c}{\textbf{NCI}} \\
\cmidrule(lr){2-3} \cmidrule(lr){4-5} \cmidrule(lr){6-7}
& \textsc{R-L} & \textsc{B-S}
& \textsc{R-L} & \textsc{B-S}
& \textsc{R-L} & \textsc{B-S}  \\
\midrule
\textsc{DeepSeek-R1} 
& \textbf{\color{rred}{14.7}} & \textbf{\color{rred}{85.1}}
& 42.0 & 90.2
& \textbf{\color{rred}{37.6}} & \textbf{\color{rred}{91.3}} \\
~$\hookrightarrow$  3-shot 
& 25.2 & 88.1
& 44.9 & 91.6
& 73.0 & 96.3 \\
~$\hookrightarrow$  5-shot 
& 25.0 & 88.0
& 44.6 & 91.8
& 75.5 & 96.6 \\
\midrule
\textsc{Kimi-K2-Inst.} 
& 18.8 & 86.7
& \textbf{\color{rred}{40.2}} & 90.3
& 68.9 & 95.6 \\
~$\hookrightarrow$  3-shot 
& \textbf{\color{OliveGreen}{27.9}} & 88.4
& \textbf{\color{OliveGreen}{52.2}} & 92.8
& 77.8 & 96.7 \\
~$\hookrightarrow$  5-shot 
& 26.4 & 88.2
& 48.3 & \textbf{\color{OliveGreen}{97.2}}
& \textbf{\color{OliveGreen}{83.7}} & \textbf{\color{OliveGreen}{97.2}} \\
\midrule
\textsc{OpenAI o3} 
& 17.3 & 86.5
& 41.5 & \textbf{\color{rred}{89.8}}
& 49.9 & 93.8 \\
~$\hookrightarrow$  3-shot 
& 26.2 & 88.5
& 51.2 & 92.6
& 71.2 & 96.0 \\
~$\hookrightarrow$  5-shot 
& 26.8 & 88.6
& 44.8 & 91.6
& 76.5 & 96.5 \\
\midrule
\textsc{GPT-5} 
& 19.2 & 86.6
& 40.6 & 89.9
& 56.4 & 93.3 \\
~$\hookrightarrow$  3-shot 
& 27.5 & \textbf{\color{OliveGreen}{88.7}}
& 46.9 & 92.3
& 70.9 & 96.4 \\
~$\hookrightarrow$  5-shot 
& 27.1 & 88.6
& 47.7 & 92.3
& 77.7 & 96.8 \\
\bottomrule
\end{tabularx}
\caption{Interpretation task results on IEI, LCI, and NCI. We report \textsc{Rouge-L} (R-L) and \textsc{BERTScore} (B-S) scores. \textbf{\textcolor{OliveGreen}{Green}} indicates the best performance and \textbf{\textcolor{rred}{red}} indicates the worst performance within each column.}
\label{tab:perf-interpretation}
\vspace{-1.5ex}
\end{table}

\paragraph{Effect of In-Context Learning (ICL).} 

The impact of ICL varies by task type (\textit{cf.} Tables~\ref{tab:major-exp-results} and \ref{tab:perf-interpretation}). Interpretation tasks benefit most consistently from few-shot prompting. Across IEI, LCI, and NCI, three- or five-shot demonstrations yield clear gains in MTR. However, as shown in Table \ref{tab:perf-interpretation}, complementary metrics reveal that these improvements primarily reflect exemplar-guided reconstruction rather than strict semantic grounding, as embedding-based similarity can be high even when lexical overlap remains limited. Few-shot ICL improves both R-L and B-S, but gains vary by phrase type, reflecting the inherently open-ended nature of interpretation outputs.

 

For classification tasks, ICL exhibits hybrid effects. Models with weaker zero-shot performance often improve, whereas others plateau or regress, such as OpenAI o3 on LCC and NCC, indicating sensitivity to exemplar selection and task formulation. Extraction tasks are the most unstable under ICL. While demonstrations can substantially improve task performance when span structure is clearly illustrated, performance may degrade when test instances diverge from the demonstrated patterns. Overall, ICL is consistently beneficial for interpretation, variably effective for classification, and highly task-dependent for extraction.


\subsection{Human Performance}
We estimate human performance using annotations from three linguistics graduate students, each labeling 100 randomly sampled examples per task in \ourbenchmark, following a two-stage protocol inspired by SuperGLUE \citep{sarlin2020superglue}. Human scores are reported as a contextual reference to situate task difficulty, rather than an upper bound on performance (\textit{cf.} Table~\ref{tab:major-exp-results}). Differences between human and model results may arise from metric properties, task ambiguity, and response normalization effects, especially for interpretation tasks. Accordingly, we avoid strong claims based on the absolute human \textit{vs.} model comparisons. Instead, human performance is used to contextualize task difficulty and to illustrate evaluation challenges under varying output constraints of \ourbenchmark.
 
\subsection{Semantic Category Scaling with ICL}
\begin{figure*}[t]
    \centering
    \begin{minipage}{1.0\linewidth}
    \includegraphics[width=\linewidth]{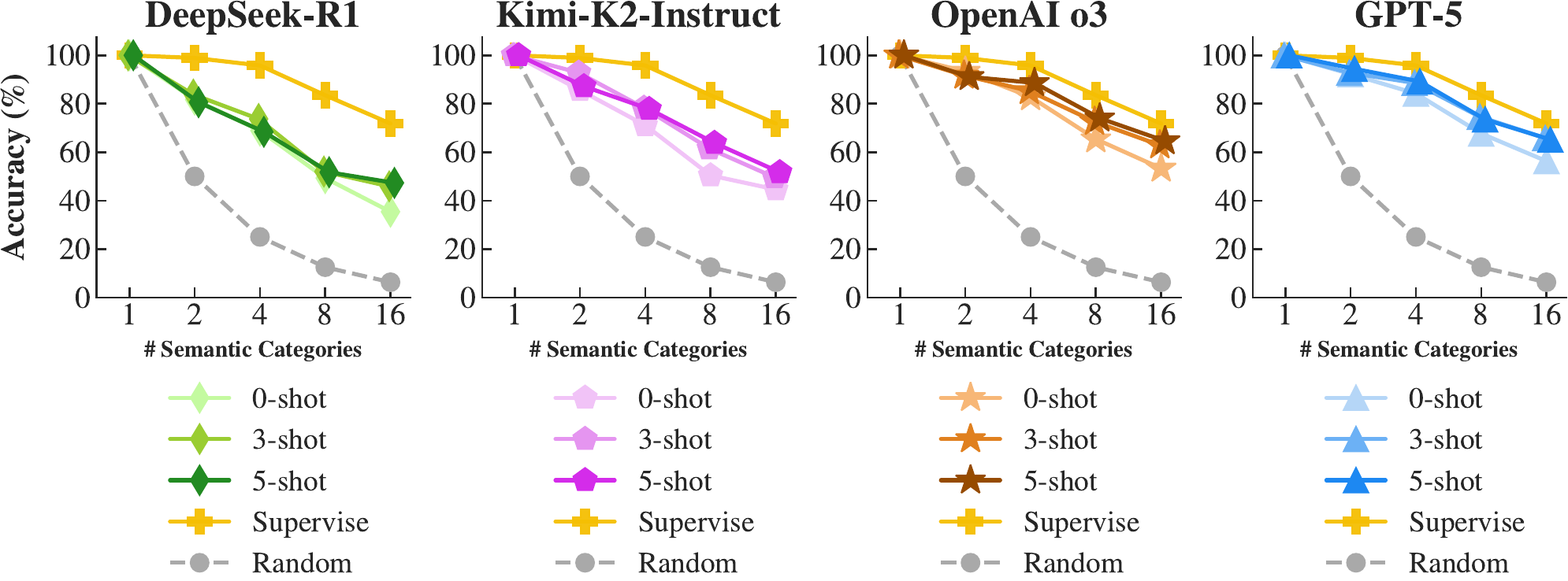}
    \end{minipage}
    \vspace{-0.5em}
    \caption{The ability of semantic relation categorization of $\mathcal{LC}$ with different numbers of in-context exemplars and semantic category scale. The number $n$ of classes is chosen from $N:= \{1, 2, 4, 8, 16\}$. Each model is prompted with the $k$-shot settings, where $k \in \{0, 3, 5\}$, respectively. Accuracy scores are calculated by the mean values based on 30 examples sampled per class from the test split of \citet{espinosa-anke-etal-2021-evaluating}, partial categories $(n \le 8)$ are run with three-class combinations in random selection, finally result in the mean value as the average.}
    \label{fig:lcc-scaling-detail}
\end{figure*}

To examine how LMs encode semantic distinctions among lexical relations, we further investigate the LCC task under an increasing number of target categories. We construct a controlled scaling setup with varying the category size from 1 to 16 by log scale, and evaluate supervised models and four representative LMs under zero- and few-shot settings. Overall results are shown in Figures~\ref{fig:lcc-scaling-detail} and \ref{fig:lcc-scaling-comparison}. Additionally, Figure~\ref{fig:llama-scaling-comparison} compares proprietary GPT-family models across all tasks, while Figure~\ref{fig:lcc-eight-classes-cm} presents the confusion matrix of GPT-5 on eight lexical function-based relations. 


Across all settings, models consistently outperform random and majority baselines, indicating non-trivial semantic reasoning even without demonstrations. Accuracy decreases as the number of categories grows, but the degradation rate varies substantially across model families. Supervised baselines remain relatively stable, whereas frontier LMs exhibit sharper drops at larger scales. For example, DeepSeek-R1 decreases from $81.7\%$ to $35.4\%$ as category size increases, suggesting that in-context semantic reasoning alone does not fully substitute for explicit supervised signals when fine-grained relational distinctions are required.

\subsection{Sequential Task Compositions} 
\label{subsec:Sequential Task Compositions}
To approximate realistic semantic phrase processing workflows, we evaluate \textbf{Sequential Task Compositions}, where models must perform multiple \textit{dependent} semantic operations in sequence, such as extraction followed by interpretation or categorization. Tables~\ref{tab:multi-step_reasoning_interpretation_results} and \ref{tab:multi-step_reasoning_judgement_results} report results for sequential interpretation and classification compositions \citep{10.24963/ijcai.2024/533, alazraki2025agentcomacompositionalbenchmarkmixing}.

For interpretation, conditional performance (Cond. MTR) on correctly extracted phrases is consistently higher than overall scores (Overall MTR) and shows only limited gains from few-shot prompting across both IE and LC settings. This gap indicates that accurate extraction remains a primary bottleneck for downstream interpretation, and that fluent semantic reconstruction does not reliably compensate for upstream structural errors.
Compositional classification degrades more sharply as task complexity increases. While leading models perform well in four-class LC settings, accuracy drops substantially in eight- and sixteen-class scenarios, with similar trends observed for IE and NC. Few-shot prompting partially mitigates this degradation but does not remove the strong dependence on extraction quality.
Overall, performance drops in compositional settings should be viewed as a diagnostic signal rather than evidence of complex error propagation. They indicate that current models struggle to robustly integrate intermediate semantic outputs, even when individual operations perform well in isolation. By separating atomic semantic operations from their compositions, \ourbenchmark\ exposes a persistent gap between performance on isolated atomic tasks and the stability of end-to-end semantic pipelines.

\begin{table}[t]
\renewcommand{\arraystretch}{0.7}
\scriptsize
\centering
\begin{tabular}{llcccc}
\toprule
\textbf{Type} & 
\textbf{Model} & 
\makecell{\textbf{\# Shot}} &
\makecell{\textbf{Ext.} \\ (\textsc{Acc})} &   
\makecell{\textbf{Cond.} \\ (\textsc{Mtr})} & 
\makecell{\textbf{Overall} \\ (\textsc{Mtr})} \\[-2.5pt]
\midrule
\multirow{8}{*}{LC}
 & \multirow{3}{*}{\textsc{DeepSeek-R1}}
 & \textit{0-shot} & 27.9 & \textcolor{rred}{\textbf{35.8}} & 10.0 \\
 &  & \textit{3-shot} & 34.4 & 38.8 & 13.4 \\
 &  & \textit{5-shot} & 33.8 & \textcolor{OliveGreen}{\textbf{42.3}} & 14.3 \\
\cmidrule(lr){2-6}
 & \multirow{3}{*}{\textsc{GPT-5}}
   & \textit{0-shot} & 26.2 & 37.6 & \textcolor{rred}{\textbf{9.9}} \\
 &  & \textit{3-shot} & 39.7 & 40.1 & 15.9 \\
 &  & \textit{5-shot} & 41.3 & 41.8 & \textcolor{OliveGreen}{\textbf{17.3}} \\
\midrule

\multirow{8}{*}{IE}
 & \multirow{3}{*}{\textsc{DeepSeek-R1}}
 & \textit{0-shot} & 51.3 & \textcolor{rred}{\textbf{12.0}} & \textcolor{rred}{\textbf{6.2}} \\
 &  & \textit{3-shot} & 57.3 & 13.0 & 7.4 \\
 &  & \textit{5-shot} & 57.0 & 13.4 & 7.6 \\
\cmidrule(lr){2-6}
 & \multirow{3}{*}{\textsc{GPT-5}}
   & \textit{0-shot} & 48.3 & \textcolor{OliveGreen}{\textbf{17.4}} & 8.4 \\
 &  & \textit{3-shot} & 55.7 & 17.2 & 9.6 \\
 &  & \textit{5-shot} & 59.3 & 17.1 & \textcolor{OliveGreen}{\textbf{10.1}} \\
\bottomrule
\end{tabular}

\caption{Performance comparison on sequential extraction-interpretation tasks. Ext. Acc denotes phrase extraction accuracy. Cond. MTR evaluates interpretation of correctly extracted phrases; Overall MTR reflects end-to-end performance. \textbf{\textcolor{OliveGreen}{Green}} indicates the highest performance across all models within each task category, while \textbf{\textcolor{rred}{red}} indicates the lowest.}
\label{tab:multi-step_reasoning_interpretation_results}
\end{table}

\begin{table}[t]
\renewcommand{\arraystretch}{1.0}  
\setlength{\tabcolsep}{2pt}      
\scriptsize
\centering
\begin{tabular}{@{}>{\centering\arraybackslash}m{0.9cm}
                >{\raggedright\arraybackslash}m{1.6cm}
                cccc cc@{}}
\toprule
\multirow{2}{*}{\textbf{\makecell{Type}}} &
\multirow{2}{*}{\textbf{Model}} &
\multicolumn{2}{c}{\textbf{0-shot}} &
\multicolumn{2}{c}{\textbf{3-shot}} &
\multicolumn{2}{c@{}}{\textbf{5-shot}} \\
\cmidrule(lr){3-4}\cmidrule(lr){5-6}\cmidrule(l){7-8}
 & & Cond. & Overall & Cond. & Overall & Cond. & Overall \\
\midrule

\multirow{8}{*}{\centering LC}
& \textsc{DeepSeek-R1} &  &  &  & &  &  \\
& ~$\hookrightarrow$ \textit{4-class} & 73.4 & 36.4 & 74.9 & 44.2 & 80.5 & 44.4 \\
& ~$\hookrightarrow$ \textit{8-class}   & 56.1 & 26.7 & 79.7 & 39.2 & 71.7 & 38.8 \\
& ~$\hookrightarrow$ \textit{16-class}   & \textcolor{rred}{\textbf{34.7}} & \textcolor{rred}{\textbf{16.0}} & 51.0 & 25.6 & 54.5 & 27.7 \\
\cmidrule(lr){2-8}
& \textsc{GPT-5} &  &  &  & &  &  \\
& ~$\hookrightarrow$ \textit{4-class}   & \textcolor{OliveGreen}{\textbf{91.3}} & 45.7 & 89.9 & \textcolor{OliveGreen}{\textbf{58.1}} & 89.9 & 55.0 \\
& ~$\hookrightarrow$ \textit{8-class}  & 76.2 & 38.8 & 83.3 & 50.0 & 80.3 & 49.2 \\
& ~$\hookrightarrow$ \textit{16-class}   & 63.4 & 33.1 & 69.4 & 43.4 & 73.4 & 44.8 \\
\midrule

\multirow{4}{*}{\centering IE}
& \textsc{DeepSeek-R1} &  &  &  & &  &  \\
& ~$\hookrightarrow$ \textit{4-class}  & 63.8 & 46.5 & 65.0 & 46.9 & \textcolor{rred}{\textbf{61.9}} & \textcolor{rred}{\textbf{45.8}} \\
& GPT-5 &  &  &  & &  &  \\
& ~$\hookrightarrow$ \textit{4-class}  & 79.3 & 65.9 & 77.7 & \textcolor{OliveGreen}{\textbf{66.3}} & \textcolor{OliveGreen}{\textbf{79.7}} & 65.9 \\
\midrule

\multirow{4}{*}{\centering NC}
& \textsc{DeepSeek-R1} &  &  &  & &  &  \\
& ~$\hookrightarrow$ \textit{4-class}  & \textcolor{rred}{\textbf{63.5}} & \textcolor{rred}{\textbf{33.2}} & \textcolor{OliveGreen}{\textbf{71.2}} & 36.9 & 71.1 & 37.8 \\
& \textsc{GPT-5} &  &  &  & &  &  \\
& ~$\hookrightarrow$ \textit{4-class}  & 68.8 & 36.5 & 64.7 & 37.3 & 66.9 & \textcolor{OliveGreen}{\textbf{38.6}} \\
\bottomrule
\end{tabular}

\caption{Classification performance comparison. Cond.: accuracy given correct extraction; Overall: end-to-end accuracy. \textbf{\textcolor{OliveGreen}{Green}} indicates the highest performance across all models (zero-shot and few-shot) within each task category, while \textbf{\textcolor{rred}{red}} indicates the lowest.}
\label{tab:multi-step_reasoning_judgement_results}
\end{table}

\begin{table}[t]
\renewcommand{\arraystretch}{1.0}
\scriptsize
\centering
\begin{tabular}{p{1.55cm}>{\centering\arraybackslash}p{0.8cm}cp{0.1em}c}
\toprule
\multirow{2.5}{*}{\textbf{Model}} & \multirow{2.5}{*}{\textbf{\# Shot}} & \multicolumn{1}{l}{\textbf{w/ \textsc{Oracle}}} && \multicolumn{1}{l}{\textbf{w/o \textsc{Oracle}}} \\ \cmidrule{3-5} && \textsc{Acc}$(\Delta)$ \raisebox{0.2ex}{$\uparrow$} && \textsc{Acc} \raisebox{0.2ex}{$\uparrow$} \\
\midrule

 \multirow{3}{*}{\textsc{Deepseek-R1}} & \textit{0-shot} & 64.1 \color{OliveGreen}{(+12.5)} & \textit{vs.} & 51.6 \\  
 & \textit{3-shot} & \textbf{72.3 \color{OliveGreen}{(+8.9)}} & \textit{vs.} & 63.4 \\ & \textit{5-shot} & 70.5 \color{OliveGreen}{(+1.2)} & \textit{vs.} & 69.3 \\
 \midrule
  \multirow{3}{*}{\textsc{Kimi-K2-Inst.}} & \textit{0-shot} & 53.3 \color{OliveGreen}{(+9.1)} & \textit{vs.} & 44.2 \\  
 & \textit{3-shot} & 67.6 \color{OliveGreen}{(+1.1)} & \textit{vs.} & 66.5 \\ & \textit{5-shot} & \textbf{69.5 \color{OliveGreen}{(+3.8)}} & \textit{vs.} & 65.7 \\ \midrule
\multirow{3}{*}{\textsc{OpenAI o3}} & \textit{0-shot} & 54.8 \color{OliveGreen}{(+6.9)} & \textit{vs.} & 47.9 \\  
 & \textit{3-shot} & 67.3 \color{OliveGreen}{(+5.1)} & \textit{vs.} & 62.2 \\ & \textit{5-shot} & \textbf{70.7 \color{OliveGreen}{(+3.6)}} & \textit{vs.} & 67.1 \\ \midrule
 
 \multirow{3}{*}{\textsc{GPT-5}} & \textit{0-shot} & 59.6 {\color{OliveGreen}{(+7.6)}} & \textit{vs.} & 52.0 \\  
 & \textit{3-shot} & 66.8 \color{OliveGreen}{(+5.1)} & \textit{vs.} & 61.7 \\ & \textit{5-shot} & \textbf{72.6 \color{OliveGreen}{(+6.9)}} & \textit{vs.} & 65.7 \\ 

\bottomrule
\end{tabular}
\caption{We report the accuracy of selected frontier LMs on the \textsc{VMwE} extraction task under different ICL setups, both with and without \textsc{Oracle Schema}.
}
\label{tab:vmwe-oracle-exp}
\end{table}

\subsection{\textsc{VMwE} Extraction with Oracle Schema}
We analyze prompting strategies for \textsc{VMwE} extraction under zero-shot and few-shot ICL settings and introduce \textsc{Oracle Schema}, which augments task instructions with the target type and its definition (\textit{cf.} Appendix \S\ref{sec:oracle-prompt-template}). Table~\ref{tab:vmwe-oracle-exp} shows that this strategy consistently improves performance across models. For example, DeepSeek-R1 increases from 51.6\% to 64.1\%, demonstrating that providing explicit semantic descriptions of the target expression substantially enhances \textsc{VMwE} extraction.

\section{Discussion and Takeaways}
Rather than restating performance trends, we distill what operation-aligned evaluation reveals about the assessment and modeling of semantics.

\paragraph{Phrasal Semantics Requires Multi-dimensional Evaluations.}
We show that phrase-level semantic competence can not be captured by any single task or metric. Interpretation, extraction, and categorization probe distinct aspects of semantic phrase processing and differ substantially in structural constraint. While extraction and categorization require explicit grounding in linguistic structure or semantic relations, interpretation operates in a weakly constrained output space. Consequently, performance on open-ended interpretation alone risks conflating fluent semantic generation with structurally grounded understanding.

\paragraph{Metric Sensitivity Shapes Apparent Model Strengths.}
The contrast between strong interpretation scores (B-S; \textit{cf.} Tables~\ref{tab:major-exp-results}, \ref{tab:perf-interpretation}, and \ref{tab:major-exp-results—appendix}) and weaker extraction performance highlights how evaluation metrics shape perceived model capabilities. 
Flexible similarity-based metrics used for interpretation primarily reward paraphrasing ability and instruction-following behavior, whereas strict span-based evaluations expose brittleness in structural identification.
As a result, high interpretation scores should be interpreted as evidence of improved exemplar-guided semantic reconstruction rather than conclusive semantic correctness. 
This discrepancy suggests that current evaluation practices may overestimate semantic robustness when structural constraints are not explicitly enforced.

\paragraph{Workflow Robustness Remains Limited.}
Sequential evaluations further reveal that semantic workflows are highly sensitive to upstream errors. Interpretation does not reliably compensate for failures in extraction or categorization; instead, structural errors propagate and often remain undetected under flexible metrics. This lack of robustness under error accumulation remains a key challenge for structured semantic evaluation settings and is largely obscured by atomic benchmarks.

\section{Conclusion}
\label{sec:conclusions}
We introduce \ourbenchmark, a benchmark for evaluating semantic phrase processing of LMs. We perform evaluations on a wide range of models with the introduced measurement, complemented by targeted human comparisons across ten tasks. The results show that, despite strong performance in general benchmarks, LMs continue to face substantial challenges in \ourbenchmark, revealing persistent limitations in semantic phrase understanding. Our analyses further characterize model behavior across task types and highlights directions for future research on more robust and structurally grounded semantic processing.

\section*{Limitations}
This work has several limitations that suggest directions for future research. First, although \ourbenchmark\ covers four common phrase phenomena, it is restricted to English and does not capture the long tail of SP types, such as multiword named entities or complex function words \citep{constant-etal-2017-survey,miletic2024semantics}. Second, while multiple task formats are included, future benchmarks should incorporate more complex sequential task compositions and additional evaluation paradigms, such as semantic retrieval \citep{espinosa-anke-etal-2021-evaluating,pham-etal-2023-pic}. 
Finally, although we evaluate many representative models, rapid progress in LM architectures calls for continual updates and broader coverage. We encourage future work to extend \ourbenchmark toward more comprehensive and multilingual resources \citep{espinosa-anke-etal-2019-collocation}.


\section*{Ethical Considerations}
This research uses publicly available datasets in accordance with their original licenses and does not include any private, sensitive, or personally identifiable information. The benchmark is intended solely for research and diagnostic purposes, and known limitations are explicitly documented to avoid overgeneralization. Computational resources were used responsibly, and potential risks related to data misuse and model evaluation were considered. Where human annotations were involved, annotators were recruited under fair labor practices and received appropriate compensation.

\section*{Acknowledgement}
\label{sec:acknowledgement}
This work was supported in part by the National Natural Science Foundation of China under Grant 62372039 and Grant 62002016, and by the Fundamental Research Funds for the Central Universities (FRF-BRA-25-012).

\bibliography{custom}

\begin{thebibliography}{88}
\providecommand{\natexlab}[1]{#1}

\bibitem[{Achiam et~al.(2023)Achiam, Adler, Agarwal, Ahmad, Akkaya, Aleman, Almeida, Altenschmidt, Altman, Anadkat et~al.}]{openai2023gpt4}
Josh Achiam, Steven Adler, Sandhini Agarwal, Lama Ahmad, Ilge Akkaya, Florencia~Leoni Aleman, Diogo Almeida, Janko Altenschmidt, Sam Altman, Shyamal Anadkat, and 1 others. 2023.
\newblock Gpt-4 technical report.
\newblock \emph{arXiv preprint arXiv:2303.08774}.

\bibitem[{Alazraki et~al.(2025)Alazraki, Chen, Brassard, Stacey, Rahmani, and Rei}]{alazraki2025agentcomacompositionalbenchmarkmixing}
Lisa Alazraki, Lihu Chen, Ana Brassard, Joe Stacey, Hossein~A. Rahmani, and Marek Rei. 2025.
\newblock \href {https://arxiv.org/abs/2508.19988} {Agentcoma: A compositional benchmark mixing commonsense and mathematical reasoning in real-world scenarios}.
\newblock \emph{Preprint}, arXiv:2508.19988.

\bibitem[{An et~al.(2025)An, Cai, Cao, Li, Lin, Liu, Lv, Ma, Wang, Wang, and Zhou}]{an2025amobenchlargelanguagemodels}
Shengnan An, Xunliang Cai, Xuezhi Cao, Xiaoyu Li, Yehao Lin, Junlin Liu, Xinxuan Lv, Dan Ma, Xuanlin Wang, Ziwen Wang, and Shuang Zhou. 2025.
\newblock \href {https://arxiv.org/abs/2510.26768} {Amo-bench: Large language models still struggle in high school math competitions}.
\newblock \emph{Preprint}, arXiv:2510.26768.

\bibitem[{Anthropic(2024)}]{TheC3}
Anthropic. 2024.
\newblock \href {https://www-cdn.anthropic.com/de8ba9b01c9ab7cbabf5c33b80b7bbc618857627/Model_Card_Claude_3.pdf} {The claude 3 model family: Opus, sonnet, haiku}.
\newblock In \emph{Anthropic Blog}.

\bibitem[{Anthropic(2025)}]{Anthropic_Claude}
Anthropic. 2025.
\newblock Anthropic.
\newblock \url{https://www.anthropic.com/news/claude-sonnet-4-5}.
\newblock September 30, 2025.

\bibitem[{Arase and Tsujii(2020)}]{arase-tsujii-2020-compositional}
Yuki Arase and Jun{'}ichi Tsujii. 2020.
\newblock \href {https://doi.org/10.18653/v1/2020.emnlp-main.125} {Compositional phrase alignment and beyond}.
\newblock In \emph{Proceedings of the 2020 Conference on Empirical Methods in Natural Language Processing (EMNLP)}, pages 1611--1623, Online. Association for Computational Linguistics.

\bibitem[{Austin et~al.(2021)Austin, Odena, Nye, Bosma, Michalewski, Dohan, Jiang, Cai, Terry, Le, and Sutton}]{Austin2021ProgramSW}
Jacob Austin, Augustus Odena, Maxwell Nye, Maarten Bosma, Henryk Michalewski, David Dohan, Ellen Jiang, Carrie~J. Cai, Michael Terry, Quoc~V. Le, and Charles Sutton. 2021.
\newblock \href {https://api.semanticscholar.org/CorpusID:237142385} {Program synthesis with large language models}.
\newblock \emph{ArXiv}, abs/2108.07732.

\bibitem[{Balunovic et~al.(2025)Balunovic, Dekoninck, Petrov, Jovanovi{\'c}, and Vechev}]{balunovic_srimatharena_2025}
Mislav Balunovic, Jasper Dekoninck, Ivo Petrov, Nikola Jovanovi{\'c}, and Martin Vechev. 2025.
\newblock \href {https://openreview.net/forum?id=y0zL9IZxZ7} {Matharena: Evaluating {LLM}s on uncontaminated math competitions}.
\newblock In \emph{The Thirty-ninth Annual Conference on Neural Information Processing Systems Datasets and Benchmarks Track}.

\bibitem[{Bisong(2019)}]{bisong2019google}
Ekaba Bisong. 2019.
\newblock Google colaboratory.
\newblock \emph{Building machine learning and deep learning models on google cloud platform: a comprehensive guide for beginners}, pages 59--64.

\bibitem[{Buijtelaar and Pezzelle(2023)}]{buijtelaar-pezzelle-2023-psycholinguistic}
Lars Buijtelaar and Sandro Pezzelle. 2023.
\newblock \href {https://doi.org/10.18653/v1/2023.eacl-main.163} {A psycholinguistic analysis of {BERT}{'}s representations of compounds}.
\newblock In \emph{Proceedings of the 17th Conference of the European Chapter of the Association for Computational Linguistics}, pages 2230--2241, Dubrovnik, Croatia. Association for Computational Linguistics.

\bibitem[{Chakrabarty et~al.(2022{\natexlab{a}})Chakrabarty, Choi, and Shwartz}]{chakrabarty-etal-2022-rocket}
Tuhin Chakrabarty, Yejin Choi, and Vered Shwartz. 2022{\natexlab{a}}.
\newblock \href {https://doi.org/10.1162/tacl_a_00478} {It{'}s not rocket science: Interpreting figurative language in narratives}.
\newblock \emph{Transactions of the Association for Computational Linguistics}, 10:589--606.

\bibitem[{Chakrabarty et~al.(2022{\natexlab{b}})Chakrabarty, Choi, and Shwartz}]{chakrabarty2022s}
Tuhin Chakrabarty, Yejin Choi, and Vered Shwartz. 2022{\natexlab{b}}.
\newblock It’s not rocket science: Interpreting figurative language in narratives.
\newblock \emph{Transactions of the Association for Computational Linguistics}, 10:589--606.

\bibitem[{Chen et~al.(2017)Chen, Long, Lu, and Huang}]{chen-etal-2017-leveraging}
I-Hsuan Chen, Yunfei Long, Qin Lu, and Chu-Ren Huang. 2017.
\newblock \href {https://doi.org/10.18653/v1/K17-1006} {Leveraging eventive information for better metaphor detection and classification}.
\newblock In \emph{Proceedings of the 21st Conference on Computational Natural Language Learning ({C}o{NLL} 2017)}, pages 36--46, Vancouver, Canada. Association for Computational Linguistics.

\bibitem[{Coil and Shwartz(2023)}]{coil-shwartz-2023-chocolate}
Albert Coil and Vered Shwartz. 2023.
\newblock \href {https://doi.org/10.18653/v1/2023.findings-acl.169} {From chocolate bunny to chocolate crocodile: Do language models understand noun compounds?}
\newblock In \emph{Findings of the Association for Computational Linguistics: ACL 2023}, pages 2698--2710, Toronto, Canada. Association for Computational Linguistics.

\bibitem[{Comanici et~al.(2025)Comanici, Bieber, Schaekermann, Pasupat, Sachdeva, Dhillon, Blistein, Ram, Zhang, Rosen et~al.}]{google_gemini}
Gheorghe Comanici, Eric Bieber, Mike Schaekermann, Ice Pasupat, Noveen Sachdeva, Inderjit Dhillon, Marcel Blistein, Ori Ram, Dan Zhang, Evan Rosen, and 1 others. 2025.
\newblock Gemini 2.5: Pushing the frontier with advanced reasoning, multimodality, long context, and next generation agentic capabilities.
\newblock \emph{arXiv preprint arXiv:2507.06261}.

\bibitem[{Constant et~al.(2017{\natexlab{a}})Constant, Eryi{\u{g}}it, Monti, Van Der~Plas, Ramisch, Rosner, and Todirascu}]{constant2017multiword}
Mathieu Constant, G{\"u}l{\c{s}}en Eryi{\u{g}}it, Johanna Monti, Lonneke Van Der~Plas, Carlos Ramisch, Michael Rosner, and Amalia Todirascu. 2017{\natexlab{a}}.
\newblock Multiword expression processing: A survey.
\newblock \emph{Computational Linguistics}, 43(4):837--892.

\bibitem[{Constant et~al.(2017{\natexlab{b}})Constant, Eryi{\v{g}}it, Monti, van~der Plas, Ramisch, Rosner, and Todirascu}]{constant-etal-2017-survey}
Mathieu Constant, G{\"u}l{\c{s}}en Eryi{\v{g}}it, Johanna Monti, Lonneke van~der Plas, Carlos Ramisch, Michael Rosner, and Amalia Todirascu. 2017{\natexlab{b}}.
\newblock \href {https://doi.org/10.1162/COLI_a_00302} {{S}urvey: Multiword expression processing: A {S}urvey}.
\newblock \emph{Computational Linguistics}, 43(4):837--892.

\bibitem[{DeepSeek(2025)}]{guo2025deepseek}
DeepSeek. 2025.
\newblock \href {https://doi.org/10.1038/s41586-025-09422-z} {Deepseek-r1 incentivizes reasoning in llms through reinforcement learning}.
\newblock \emph{Nature}, 645:633--638.

\bibitem[{Denkowski and Lavie(2014)}]{denkowski-lavie-2014-meteor}
Michael Denkowski and Alon Lavie. 2014.
\newblock \href {https://doi.org/10.3115/v1/W14-3348} {Meteor universal: Language specific translation evaluation for any target language}.
\newblock In \emph{Proceedings of the Ninth Workshop on Statistical Machine Translation}, pages 376--380, Baltimore, Maryland, USA. Association for Computational Linguistics.

\bibitem[{Devlin et~al.(2019)Devlin, Chang, Lee, and Toutanova}]{devlin2019bert}
Jacob Devlin, Ming-Wei Chang, Kenton Lee, and Kristina Toutanova. 2019.
\newblock Bert: Pre-training of deep bidirectional transformers for language understanding.
\newblock In \emph{Proceedings of the 2019 Conference of the North American Chapter of the Association for Computational Linguistics: Human Language Technologies, Volume 1 (Long and Short Papers)}, pages 4171--4186.

\bibitem[{Espinosa-Anke et~al.(2021)Espinosa-Anke, Codina-Filba, and Wanner}]{espinosa-anke-etal-2021-evaluating}
Luis Espinosa-Anke, Joan Codina-Filba, and Leo Wanner. 2021.
\newblock \href {https://doi.org/10.18653/v1/2021.eacl-main.120} {Evaluating language models for the retrieval and categorization of lexical collocations}.
\newblock In \emph{Proceedings of the 16th Conference of the European Chapter of the Association for Computational Linguistics: Main Volume}, pages 1406--1417, Online. Association for Computational Linguistics.

\bibitem[{Espinosa-Anke et~al.(2019)Espinosa-Anke, Schockaert, and Wanner}]{espinosa-anke-etal-2019-collocation}
Luis Espinosa-Anke, Steven Schockaert, and Leo Wanner. 2019.
\newblock \href {https://doi.org/10.18653/v1/P19-1576} {Collocation classification with unsupervised relation vectors}.
\newblock In \emph{Proceedings of the 57th Annual Meeting of the Association for Computational Linguistics}, pages 5765--5772, Florence, Italy. Association for Computational Linguistics.

\bibitem[{Espinosa-Anke et~al.(2022)Espinosa-Anke, Shvets, Mohammadshahi, Henderson, and Wanner}]{espinosa-anke-etal-2022-multilingual}
Luis Espinosa-Anke, Alexander Shvets, Alireza Mohammadshahi, James Henderson, and Leo Wanner. 2022.
\newblock \href {https://doi.org/10.18653/v1/2022.starsem-1.8} {Multilingual extraction and categorization of lexical collocations with graph-aware transformers}.
\newblock In \emph{Proceedings of the 11th Joint Conference on Lexical and Computational Semantics}, pages 89--100, Seattle, Washington. Association for Computational Linguistics.

\bibitem[{Fazly et~al.(2009)Fazly, Cook, and Stevenson}]{fazly2009unsupervised}
Afsaneh Fazly, Paul Cook, and Suzanne Stevenson. 2009.
\newblock Unsupervised type and token identification of idiomatic expressions.
\newblock \emph{Computational Linguistics}, 35(1):61--103.

\bibitem[{Fisas et~al.(2020)Fisas, Espinosa-Anke, Codina-Filb{\'a}, and Wanner}]{fisas-etal-2020-collfren}
Beatriz Fisas, Luis Espinosa-Anke, Joan Codina-Filb{\'a}, and Leo Wanner. 2020.
\newblock \href {https://aclanthology.org/2020.mwe-1.1} {{C}oll{F}r{E}n: Rich bilingual {E}nglish{--}{F}rench collocation resource}.
\newblock In \emph{Proceedings of the Joint Workshop on Multiword Expressions and Electronic Lexicons}, pages 1--12, online. Association for Computational Linguistics.

\bibitem[{Garcia et~al.(2021)Garcia, Kramer~Vieira, Scarton, Idiart, and Villavicencio}]{garcia-etal-2021-assessing}
Marcos Garcia, Tiago Kramer~Vieira, Carolina Scarton, Marco Idiart, and Aline Villavicencio. 2021.
\newblock \href {https://doi.org/10.18653/v1/2021.acl-long.212} {Assessing the representations of idiomaticity in vector models with a noun compound dataset labeled at type and token levels}.
\newblock In \emph{Proceedings of the 59th Annual Meeting of the Association for Computational Linguistics and the 11th International Joint Conference on Natural Language Processing (Volume 1: Long Papers)}, pages 2730--2741, Online. Association for Computational Linguistics.

\bibitem[{Gelbukh et~al.(2012)}]{gelbukh2012semantic}
Alexander Gelbukh and 1 others. 2012.
\newblock \emph{Semantic analysis of verbal collocations with lexical functions}, volume 414.
\newblock Springer.

\bibitem[{Harish et~al.(2021)Harish, Edward, Carolina, and Aline}]{tayyar-madabushi-etal-2021-astitchinlanguagemodels-dataset}
Tayyar~{Madabushi} Harish, Gow-Smith Edward, Scarton Carolina, and Villavicencio Aline. 2021.
\newblock \href {https://doi.org/10.18653/v1/2021.findings-emnlp.294} {{AS}titch{I}n{L}anguage{M}odels: Dataset and methods for the exploration of idiomaticity in pre-trained language models}.
\newblock In \emph{Findings of the Association for Computational Linguistics: EMNLP 2021}, pages 3464--3477, Punta Cana, Dominican Republic. Association for Computational Linguistics.

\bibitem[{Haviv et~al.(2023)Haviv, Cohen, Gidron, Schuster, Goldberg, and Geva}]{haviv-etal-2023-understanding}
Adi Haviv, Ido Cohen, Jacob Gidron, Roei Schuster, Yoav Goldberg, and Mor Geva. 2023.
\newblock \href {https://doi.org/10.18653/v1/2023.eacl-main.19} {Understanding transformer memorization recall through idioms}.
\newblock In \emph{Proceedings of the 17th Conference of the European Chapter of the Association for Computational Linguistics}, pages 248--264, Dubrovnik, Croatia. Association for Computational Linguistics.

\bibitem[{Hendrickx et~al.(2013)Hendrickx, Kozareva, Nakov, {\'O}~S{\'e}aghdha, Szpakowicz, and Veale}]{hendrickx-etal-2013-semeval}
Iris Hendrickx, Zornitsa Kozareva, Preslav Nakov, Diarmuid {\'O}~S{\'e}aghdha, Stan Szpakowicz, and Tony Veale. 2013.
\newblock \href {https://aclanthology.org/S13-2025} {{S}em{E}val-2013 task 4: Free paraphrases of noun compounds}.
\newblock In \emph{Second Joint Conference on Lexical and Computational Semantics (*{SEM}), Volume 2: Proceedings of the Seventh International Workshop on Semantic Evaluation ({S}em{E}val 2013)}, pages 138--143, Atlanta, Georgia, USA. Association for Computational Linguistics.

\bibitem[{Holtzman et~al.(2019)Holtzman, Buys, Du, Forbes, and Choi}]{holtzman2019curious}
Ari Holtzman, Jan Buys, Li~Du, Maxwell Forbes, and Yejin Choi. 2019.
\newblock The curious case of neural text degeneration.
\newblock In \emph{International Conference on Learning Representations}.

\bibitem[{Hu et~al.(2024)Hu, Chen, Li, Guo, Wen, Yu, and Guo}]{ICLR2024_7b7d7985}
Xuming Hu, Junzhe Chen, Xiaochuan Li, Yufei Guo, Lijie Wen, Philip~S. Yu, and Zhijiang Guo. 2024.
\newblock \href {https://openreview.net/forum?id=9OevMUdods} {Towards understanding factual knowledge of large language models}.
\newblock In \emph{The Twelfth International Conference on Learning Representations}.

\bibitem[{Huang et~al.(2025)Huang, Gu, Li, Hu, Qing, and Xu}]{huang-etal-2025-structfact}
Sirui Huang, Yanggan Gu, Zhonghao Li, Xuming Hu, Li~Qing, and Guandong Xu. 2025.
\newblock \href {https://doi.org/10.18653/v1/2025.findings-acl.391} {{S}truct{F}act: Reasoning factual knowledge from structured data with large language models}.
\newblock In \emph{Findings of the Association for Computational Linguistics: ACL 2025}, pages 7521--7552, Vienna, Austria. Association for Computational Linguistics.

\bibitem[{Kamath et~al.(2025)Kamath, Ferret, Pathak, Vieillard, Merhej, Perrin, Matejovicova, Ram'e, Rivi{\`e}re, Rouillard, Mesnard, Cideron, Grill, Ramos, Yvinec, Casbon, Pot, Penchev, Liu, Visin, Kenealy, Beyer, Zhai, Tsitsulin, Busa-Fekete, Feng, Sachdeva, Coleman, Gao, Mustafa, Barr, Parisotto, Tian, Eyal, Cherry, Peter, Sinopalnikov, Bhupatiraju, Agarwal, Kazemi, Malkin, Kumar, Vilar, Brusilovsky, Luo, Steiner, Friesen, Sharma, Sharma, Gilady, Goedeckemeyer, Saade, Kolesnikov, Bendebury, Abdagic, Vadi, Gyorgy, Pinto, Das, Bapna, Miech, Yang, Paterson, Shenoy, Chakrabarti, Piot, Wu, Shahriari, Petrini, Chen, Lan, Choquette-Choo, Carey, Brick, Deutsch, Eisenbud, Cattle, Cheng, Paparas, Sreepathihalli, Reid, Tran, Zelle, Noland, Huizenga, Kharitonov, Liu, Amirkhanyan, Cameron, Hashemi, Klimczak-Pluci'nska, Singh, Mehta, Lehri, Hazimeh, Ballantyne, Szpektor, Nardini, Pouget-Abadie, Chan, Stanton, Wieting, Lai, Orbay, Fernandez, Newlan, Ji, Singh, Black, Yu, Hui, Vodrahalli, Greff, Qiu, Valentine, Coelho,
  Ritter, Hoffman, Watson, Chaturvedi, Moynihan, Ma, Babar, Noy, Byrd, Roy, Momchev, Chauhan, Bunyan, Botarda, Caron, Rubenstein, Culliton, Schmid, Sessa, mei Xu, Stańczyk, Tafti, Shivanna, Wu, Pan, Rokni, Willoughby, Vallu, Mullins, Jerome, Smoot, Girgin, Iqbal, Reddy, Sheth, P{\~o}der, Bhatnagar, Panyam, Eiger, Zhang, Liu, Yacovone, Liechty, Kalra, Evci, Misra, Roseberry, Feinberg, Kolesnikov, Han, Kwon, Chen, Chow, Zhu, Wei, Egyed, Cotruta, Giang, Kirk, Rao, Lo, Moreira, Martins, Sanseviero, Gonzalez, Gleicher, Warkentin, Mirrokni, Senter, Collins, Barral, Ghahramani, Hadsell, Matias, Sculley, Petrov, Fiedel, Shazeer, Vinyals, Dean, Hassabis, Kavukcuoglu, Farabet, Buchatskaya, Alayrac, Anil, Lepikhin, Borgeaud, Bachem, Joulin, Andreev, Hardin, Dadashi, and Hussenot}]{gemma_2025}
Gemma Team~Aishwarya Kamath, Johan Ferret, Shreya Pathak, Nino Vieillard, Ramona Merhej, Sarah Perrin, Tatiana Matejovicova, Alexandre Ram'e, Morgane Rivi{\`e}re, Louis Rouillard, Thomas Mesnard, Geoffrey Cideron, Jean-Bastien Grill, Sabela Ramos, Edouard Yvinec, Michelle Casbon, Etienne Pot, Ivo Penchev, Gael Liu, and 191 others. 2025.
\newblock \href {https://api.semanticscholar.org/CorpusID:277313563} {Gemma 3 technical report}.
\newblock \emph{ArXiv}, abs/2503.19786.

\bibitem[{{Kimi}(2025)}]{kimiteam2025kimik2openagentic}
{Kimi}. 2025.
\newblock \href {https://arxiv.org/abs/2507.20534} {Kimi k2: Open agentic intelligence}.
\newblock \emph{Preprint}, arXiv:2507.20534.

\bibitem[{Klubi{\v{c}}ka et~al.(2023)Klubi{\v{c}}ka, Nedumpozhimana, and Kelleher}]{klubicka-etal-2023-idioms}
Filip Klubi{\v{c}}ka, Vasudevan Nedumpozhimana, and John Kelleher. 2023.
\newblock \href {https://doi.org/10.18653/v1/2023.mwe-1.8} {Idioms, probing and dangerous things: Towards structural probing for idiomaticity in vector space}.
\newblock In \emph{Proceedings of the 19th Workshop on Multiword Expressions (MWE 2023)}, pages 45--57, Dubrovnik, Croatia. Association for Computational Linguistics.

\bibitem[{Kolesnikova(2020)}]{kolesnikova2020automatic}
Olga Kolesnikova. 2020.
\newblock Automatic detection of lexical functions in context.
\newblock \emph{Computaci{\'o}n y sistemas}, 24(3):1337--1352.

\bibitem[{Kolluru et~al.(2022)Kolluru, Stanovsky, and {Mausam}}]{kolluru-etal-2022-covid}
Keshav Kolluru, Gabriel Stanovsky, and {Mausam}. 2022.
\newblock \href {https://doi.org/10.18653/v1/2022.emnlp-main.711} {{``}covid vaccine is against covid but {O}xford vaccine is made at {O}xford!{''} semantic interpretation of proper noun compounds}.
\newblock In \emph{Proceedings of the 2022 Conference on Empirical Methods in Natural Language Processing}, pages 10407--10420, Abu Dhabi, United Arab Emirates. Association for Computational Linguistics.

\bibitem[{Kwon et~al.(2023)Kwon, Li, Zhuang, Sheng, Zheng, Yu, Gonzalez, Zhang, and Stoica}]{kwon2023efficient}
Woosuk Kwon, Zhuohan Li, Siyuan Zhuang, Ying Sheng, Lianmin Zheng, Cody~Hao Yu, Joseph~E. Gonzalez, Hao Zhang, and Ion Stoica. 2023.
\newblock Efficient memory management for large language model serving with pagedattention.
\newblock In \emph{Proceedings of the ACM SIGOPS 29th Symposium on Operating Systems Principles}.

\bibitem[{Li et~al.(2024)Li, Li, Zhang, Zhao, Dong, Jin, Li, Huang, and Li}]{NEURIPS2024_6a059625}
Jia Li, Ge~Li, Xuanming Zhang, Yunfei Zhao, Yihong Dong, Zhi Jin, Binhua Li, Fei Huang, and Yongbin Li. 2024.
\newblock \href {https://doi.org/10.52202/079017-1837} {Evocodebench: An evolving code generation benchmark with domain-specific evaluations}.
\newblock In \emph{Advances in Neural Information Processing Systems}, volume~37, pages 57619--57641. Curran Associates, Inc.

\bibitem[{Li et~al.(2025)Li, Dong, Liu, Yang, Wang, Wang, Zhu, Jia, and Zheng}]{li2025reflectevo}
Jiaqi Li, Xinyi Dong, Yang Liu, Zhizhuo Yang, Quansen Wang, Xiaobo Wang, Song-Chun Zhu, Zixia Jia, and Zilong Zheng. 2025.
\newblock \href {https://doi.org/10.18653/v1/2025.findings-acl.871} {{R}eflect{E}vo: Improving meta introspection of small {LLM}s by learning self-reflection}.
\newblock In \emph{Findings of the Association for Computational Linguistics: ACL 2025}, pages 16948--16966, Vienna, Austria. Association for Computational Linguistics.

\bibitem[{Lin(2004)}]{lin-2004-rouge}
Chin-Yew Lin. 2004.
\newblock \href {https://aclanthology.org/W04-1013} {{ROUGE}: A package for automatic evaluation of summaries}.
\newblock In \emph{Text Summarization Branches Out}, pages 74--81, Barcelona, Spain. Association for Computational Linguistics.

\bibitem[{Liu et~al.(2024)Liu, Zheng, Qiao, Duan, Fei, Zhou, Zhang, Zhang, Lin, and Chen}]{liu2024mathbenchevaluatingtheoryapplication}
Hongwei Liu, Zilong Zheng, Yuxuan Qiao, Haodong Duan, Zhiwei Fei, Fengzhe Zhou, Wenwei Zhang, Songyang Zhang, Dahua Lin, and Kai Chen. 2024.
\newblock \href {https://doi.org/10.18653/v1/2024.findings-acl.411} {{M}ath{B}ench: Evaluating the theory and application proficiency of {LLM}s with a hierarchical mathematics benchmark}.
\newblock In \emph{Findings of the Association for Computational Linguistics: ACL 2024}, pages 6884--6915, Bangkok, Thailand. Association for Computational Linguistics.

\bibitem[{Liu et~al.(2026{\natexlab{a}})Liu, Li, and Zheng}]{liu2026rulereasoner}
Yang Liu, Jiaqi Li, and Zilong Zheng. 2026{\natexlab{a}}.
\newblock \href {https://openreview.net/forum?id=MQV4TJyqnb} {Rulereasoner: Reinforced rule-based reasoning via domain-aware dynamic sampling}.
\newblock In \emph{The Fourteenth International Conference on Learning Representations}.

\bibitem[{Liu et~al.(2026{\natexlab{b}})Liu, Yang, Li, Liang, Li, and Yan}]{liu-etal-2026-lm}
Yang Liu, Jiaye Yang, Weikang Li, Jiahui Liang, Yang Li, and Lingyong Yan. 2026{\natexlab{b}}.
\newblock \href {https://doi.org/10.18653/v1/2026.eacl-long.1} {{LM}-lexicon: Improving definition modeling via harmonizing semantic experts}.
\newblock In \emph{Proceedings of the 19th Conference of the {E}uropean Chapter of the {A}ssociation for {C}omputational {L}inguistics (Volume 1: Long Papers)}, pages 1--22, Rabat, Morocco. Association for Computational Linguistics.

\bibitem[{Luong et~al.(2025)Luong, Hwang, Nguyen, Ghiasi, Chervonyi, Seo, Kim, Bingham, Lee, Mishra, Zhai, Hu, Michalewski, Kim, Ahn, Bae, Song, Trinh, Le, and Jung}]{luong-etal-2025-towards}
Thang Luong, Dawsen Hwang, Hoang~H Nguyen, Golnaz Ghiasi, Yuri Chervonyi, Insuk Seo, Junsu Kim, Garrett Bingham, Jonathan Lee, Swaroop Mishra, Alex Zhai, Huiyi Hu, Henryk Michalewski, Jimin Kim, Jeonghyun Ahn, Junhwi Bae, Xingyou Song, Trieu~Hoang Trinh, Quoc~V Le, and Junehyuk Jung. 2025.
\newblock \href {https://doi.org/10.18653/v1/2025.emnlp-main.1794} {Towards robust mathematical reasoning}.
\newblock In \emph{Proceedings of the 2025 Conference on Empirical Methods in Natural Language Processing}, pages 35406--35430, Suzhou, China. Association for Computational Linguistics.

\bibitem[{Mel'\v{c}uk(1998)}]{mel1998collocations}
Igor~A. Mel'\v{c}uk. 1998.
\newblock Collocations and lexical functions.
\newblock \emph{Phraseology. Theory, analysis, and applications}, pages 23--53.

\bibitem[{Mel'\v{c}uk(2023)}]{mel2023general}
Igor~A. Mel'\v{c}uk. 2023.
\newblock \emph{General phraseology: Theory and practice}.
\newblock John Benjamins.

\bibitem[{Mileti{\'c} and Schulte~im Walde(2024)}]{miletic2024semantics}
Filip Mileti{\'c} and Sabine Schulte~im Walde. 2024.
\newblock \href {https://doi.org/10.1162/tacl_a_00657} {Semantics of multiword expressions in transformer-based models: A survey}.
\newblock \emph{Transactions of the Association for Computational Linguistics}, 12:593--612.

\bibitem[{Nunberg et~al.(1994)Nunberg, Sag, and Wasow}]{54311d36-f6a7-378f-b3bb-f2b37bdfb9f1}
Geoffrey Nunberg, Ivan~A. Sag, and Thomas Wasow. 1994.
\newblock \href {http://www.jstor.org/stable/416483} {Idioms}.
\newblock \emph{Language}, 70(3):491--538.

\bibitem[{OpenAI(2025)}]{openai_o3}
OpenAI. 2025.
\newblock Openai.
\newblock \url{https://openai.com/index/introducing-o3-and-o4-mini}.
\newblock April 16, 2025.

\bibitem[{Pasquer et~al.(2020)Pasquer, Savary, Ramisch, and Antoine}]{pasquer-etal-2020-verbal}
Caroline Pasquer, Agata Savary, Carlos Ramisch, and Jean-Yves Antoine. 2020.
\newblock \href {https://doi.org/10.18653/v1/2020.coling-main.296} {Verbal multiword expression identification: Do we need a sledgehammer to crack a nut?}
\newblock In \emph{Proceedings of the 28th International Conference on Computational Linguistics}, pages 3333--3345, Barcelona, Spain (Online). International Committee on Computational Linguistics.

\bibitem[{Paszke et~al.(2019)Paszke, Gross, Massa, Lerer, Bradbury, Chanan, Killeen, Lin, Gimelshein, Antiga et~al.}]{paszke2019pytorch}
Adam Paszke, Sam Gross, Francisco Massa, Adam Lerer, James Bradbury, Gregory Chanan, Trevor Killeen, Zeming Lin, Natalia Gimelshein, Luca Antiga, and 1 others. 2019.
\newblock Pytorch: An imperative style, high-performance deep learning library.
\newblock \emph{Advances in neural information processing systems}, 32.

\bibitem[{Pham et~al.(2023)Pham, Yoon, Bui, and Nguyen}]{pham-etal-2023-pic}
Thang Pham, Seunghyun Yoon, Trung Bui, and Anh Nguyen. 2023.
\newblock \href {https://doi.org/10.18653/v1/2023.eacl-main.1} {{P}i{C}: A phrase-in-context dataset for phrase understanding and semantic search}.
\newblock In \emph{Proceedings of the 17th Conference of the European Chapter of the Association for Computational Linguistics}, pages 1--26, Dubrovnik, Croatia. Association for Computational Linguistics.

\bibitem[{Radford et~al.(2019)Radford, Wu, Child, Luan, Amodei, and Sutskever}]{radford2019language}
Alec Radford, Jeffrey Wu, Rewon Child, David Luan, Dario Amodei, and Ilya Sutskever. 2019.
\newblock \href {https://cdn.openai.com/better-language-models/language_models_are_unsupervised_multitask_learners.pdf} {Language models are unsupervised multitask learners}.
\newblock \emph{OpenAI}.
\newblock Accessed: 2024-11-15.

\bibitem[{Raffel et~al.(2020)Raffel, Shazeer, Roberts, Lee, Narang, Matena, Zhou, Li, and Liu}]{raffel2020exploring}
Colin Raffel, Noam Shazeer, Adam Roberts, Katherine Lee, Sharan Narang, Michael Matena, Yanqi Zhou, Wei Li, and Peter~J Liu. 2020.
\newblock Exploring the limits of transfer learning with a unified text-to-text transformer.
\newblock \emph{The Journal of Machine Learning Research}, 21(1):5485--5551.

\bibitem[{Ram et~al.(2024)Ram, Klinger, and Gray}]{10.24963/ijcai.2024/533}
Parikshit Ram, Tim Klinger, and Alexander~G. Gray. 2024.
\newblock \href {https://doi.org/10.24963/ijcai.2024/533} {What makes models compositional? a theoretical view}.
\newblock In \emph{Proceedings of the Thirty-Third International Joint Conference on Artificial Intelligence}, IJCAI '24.

\bibitem[{Ramisch(2023)}]{ramisch2023multiword}
Carlos Ramisch. 2023.
\newblock \href {https://theses.hal.science/tel-04216223} {\emph{{Multiword expressions in computational linguistics}}}.
\newblock Habilitation {\`a} diriger des recherches, {Aix Marseille Universit{\'e} (AMU)}.

\bibitem[{Ramisch et~al.(2020)Ramisch, Savary, Guillaume, Waszczuk, Candito, Vaidya, Barbu~Mititelu, Bhatia, I{\~n}urrieta, Giouli, G{\"u}ng{\"o}r, Jiang, Lichte, Liebeskind, Monti, Ramisch, Stymne, Walsh, and Xu}]{ramisch-etal-2020-edition}
Carlos Ramisch, Agata Savary, Bruno Guillaume, Jakub Waszczuk, Marie Candito, Ashwini Vaidya, Verginica Barbu~Mititelu, Archna Bhatia, Uxoa I{\~n}urrieta, Voula Giouli, Tunga G{\"u}ng{\"o}r, Menghan Jiang, Timm Lichte, Chaya Liebeskind, Johanna Monti, Renata Ramisch, Sara Stymne, Abigail Walsh, and Hongzhi Xu. 2020.
\newblock \href {https://aclanthology.org/2020.mwe-1.14} {Edition 1.2 of the {PARSEME} shared task on semi-supervised identification of verbal multiword expressions}.
\newblock In \emph{Proceedings of the Joint Workshop on Multiword Expressions and Electronic Lexicons}, pages 107--118, online. Association for Computational Linguistics.

\bibitem[{Ramisch et~al.(2023{\natexlab{a}})Ramisch, Walsh, Blanchard, and Taslimipoor}]{ramisch2023survey}
Carlos Ramisch, Abigail Walsh, Thomas Blanchard, and Shiva Taslimipoor. 2023{\natexlab{a}}.
\newblock A survey of mwe identification experiments: The devil is in the details.
\newblock In \emph{Proceedings of the 19th Workshop on Multiword Expressions (MWE 2023)}, pages 106--120.

\bibitem[{Ramisch et~al.(2023{\natexlab{b}})Ramisch, Walsh, Blanchard, and Taslimipoor}]{ramisch-etal-2023-survey}
Carlos Ramisch, Abigail Walsh, Thomas Blanchard, and Shiva Taslimipoor. 2023{\natexlab{b}}.
\newblock \href {https://doi.org/10.18653/v1/2023.mwe-1.15} {A survey of {MWE} identification experiments: The devil is in the details}.
\newblock In \emph{Proceedings of the 19th Workshop on Multiword Expressions (MWE 2023)}, pages 106--120, Dubrovnik, Croatia. Association for Computational Linguistics.

\bibitem[{Rodr{\'\i}guez(2003)}]{rodriguez2003domain}
Mar{\'\i}a A~Barrios Rodr{\'\i}guez. 2003.
\newblock The domain of the lexical functions fact0, causfact0 and real1.
\newblock \emph{learning}, page~64.

\bibitem[{Sag et~al.(2002{\natexlab{a}})Sag, Baldwin, Bond, Copestake, and Flickinger}]{sag2002multiword}
Ivan~A Sag, Timothy Baldwin, Francis Bond, Ann Copestake, and Dan Flickinger. 2002{\natexlab{a}}.
\newblock Multiword expressions: A pain in the neck for nlp.
\newblock In \emph{Computational Linguistics and Intelligent Text Processing: Third International Conference, CICLing 2002 Mexico City, Mexico, February 17--23, 2002 Proceedings 3}, pages 1--15. Springer.

\bibitem[{Sag et~al.(2002{\natexlab{b}})Sag, Baldwin, Bond, Copestake, and Flickinger}]{10.1007/3-540-45715-1_1}
Ivan~A. Sag, Timothy Baldwin, Francis Bond, Ann Copestake, and Dan Flickinger. 2002{\natexlab{b}}.
\newblock Multiword expressions: A pain in the neck for nlp.
\newblock In \emph{Computational Linguistics and Intelligent Text Processing}, pages 1--15, Berlin, Heidelberg. Springer Berlin Heidelberg.

\bibitem[{Sailer and Markantonatou(2018)}]{sailer2018multiword}
Manfred Sailer and Stella Markantonatou. 2018.
\newblock \emph{Multiword expressions: Insights from a multi-lingual perspective}.
\newblock Language Science Press.

\bibitem[{Sarlin et~al.(2020)Sarlin, DeTone, Malisiewicz, and Rabinovich}]{sarlin2020superglue}
Paul-Edouard Sarlin, Daniel DeTone, Tomasz Malisiewicz, and Andrew Rabinovich. 2020.
\newblock Superglue: Learning feature matching with graph neural networks.
\newblock In \emph{Proceedings of the IEEE/CVF conference on computer vision and pattern recognition}, pages 4938--4947.

\bibitem[{Savary et~al.(2023)Savary, Ben~Khelil, Ramisch, Giouli, Barbu~Mititelu, Hadj~Mohamed, Krstev, Liebeskind, Xu, Stymne, G{\"u}ng{\"o}r, Pickard, Guillaume, Bej{\v{c}}ek, Bhatia, Candito, Gantar, I{\~n}urrieta, Gatt, Kovalevskaite, Lichte, Ljube{\v{s}}i{\'c}, Monti, Parra~Escart{\'\i}n, Shamsfard, Stoyanova, Vincze, and Walsh}]{savary-etal-2023-parseme}
Agata Savary, Cherifa Ben~Khelil, Carlos Ramisch, Voula Giouli, Verginica Barbu~Mititelu, Najet Hadj~Mohamed, Cvetana Krstev, Chaya Liebeskind, Hongzhi Xu, Sara Stymne, Tunga G{\"u}ng{\"o}r, Thomas Pickard, Bruno Guillaume, Eduard Bej{\v{c}}ek, Archna Bhatia, Marie Candito, Polona Gantar, Uxoa I{\~n}urrieta, Albert Gatt, and 9 others. 2023.
\newblock \href {https://doi.org/10.18653/v1/2023.mwe-1.6} {{PARSEME} corpus release 1.3}.
\newblock In \emph{Proceedings of the 19th Workshop on Multiword Expressions (MWE 2023)}, pages 24--35, Dubrovnik, Croatia. Association for Computational Linguistics.

\bibitem[{Savary et~al.(2017)Savary, Ramisch, Cordeiro, Sangati, Vincze, QasemiZadeh, Candito, Cap, Giouli, Stoyanova, and Doucet}]{savary-etal-2017-parseme}
Agata Savary, Carlos Ramisch, Silvio Cordeiro, Federico Sangati, Veronika Vincze, Behrang QasemiZadeh, Marie Candito, Fabienne Cap, Voula Giouli, Ivelina Stoyanova, and Antoine Doucet. 2017.
\newblock \href {https://doi.org/10.18653/v1/W17-1704} {The {PARSEME} shared task on automatic identification of verbal multiword expressions}.
\newblock In \emph{Proceedings of the 13th Workshop on Multiword Expressions ({MWE} 2017)}, pages 31--47, Valencia, Spain. Association for Computational Linguistics.

\bibitem[{Shvets and Wanner(2022)}]{math10203831}
Alexander Shvets and Leo Wanner. 2022.
\newblock \href {https://doi.org/10.3390/math10203831} {The relation dimension in the identification and classification of lexically restricted word co-occurrences in text corpora}.
\newblock \emph{Mathematics}, 10(20).

\bibitem[{Shwartz and Dagan(2019)}]{shwartz-dagan-2019-still}
Vered Shwartz and Ido Dagan. 2019.
\newblock \href {https://doi.org/10.1162/tacl_a_00277} {Still a pain in the neck: Evaluating text representations on lexical composition}.
\newblock \emph{Transactions of the Association for Computational Linguistics}, 7:403--419.

\bibitem[{Singh et~al.(2025)Singh, Fry, Perelman, Tart, Ganesh, El-Kishky, McLaughlin, Low, Ostrow, Ananthram et~al.}]{openai_gpt5}
Aaditya Singh, Adam Fry, Adam Perelman, Adam Tart, Adi Ganesh, Ahmed El-Kishky, Aidan McLaughlin, Aiden Low, AJ~Ostrow, Akhila Ananthram, and 1 others. 2025.
\newblock Openai gpt-5 system card.
\newblock \emph{arXiv preprint arXiv:2601.03267}.

\bibitem[{Spathas and Michelioudakis(2021)}]{Spathas2021}
Giorgos Spathas and Dimitris Michelioudakis. 2021.
\newblock \href {https://doi.org/10.1007/s11049-020-09496-6} {States in the decomposition of verbal predicates}.
\newblock \emph{Natural Language \& Linguistic Theory}, 39(4):1253--1306.

\bibitem[{Tanner and Hoffman(2023)}]{tanner-hoffman-2023-mwe}
Joshua Tanner and Jacob Hoffman. 2023.
\newblock \href {https://doi.org/10.18653/v1/2023.findings-emnlp.14} {{MWE} as {WSD}: Solving multiword expression identification with word sense disambiguation}.
\newblock In \emph{Findings of the Association for Computational Linguistics: EMNLP 2023}, pages 181--193, Singapore. Association for Computational Linguistics.

\bibitem[{Tedeschi et~al.(2022)Tedeschi, Martelli, and Navigli}]{tedeschi-etal-2022-id10m}
Simone Tedeschi, Federico Martelli, and Roberto Navigli. 2022.
\newblock \href {https://doi.org/10.18653/v1/2022.findings-naacl.208} {{ID}10{M}: Idiom identification in 10 languages}.
\newblock In \emph{Findings of the Association for Computational Linguistics: NAACL 2022}, pages 2715--2726, Seattle, United States. Association for Computational Linguistics.

\bibitem[{Thielen(1999)}]{fontenelle1997turning}
Christine Thielen. 1999.
\newblock Review of "turning a bilingual dictionary into a lexical-semantic database" by thierry fontenelle. max niemeyer verlag 1997.
\newblock \emph{Comput. Linguist.}, 25(3):447–449.

\bibitem[{Tratz and Hovy(2010)}]{tratz-hovy-2010-taxonomy}
Stephen Tratz and Eduard Hovy. 2010.
\newblock \href {https://aclanthology.org/P10-1070/} {A taxonomy, dataset, and classifier for automatic noun compound interpretation}.
\newblock In \emph{Proceedings of the 48th Annual Meeting of the Association for Computational Linguistics}, pages 678--687, Uppsala, Sweden. Association for Computational Linguistics.

\bibitem[{Vacareanu et~al.(2020)Vacareanu, Valenzuela-Esc{\'a}rcega, Sharp, and Surdeanu}]{vacareanu-etal-2020-unsupervised}
Robert Vacareanu, Marco~A. Valenzuela-Esc{\'a}rcega, Rebecca Sharp, and Mihai Surdeanu. 2020.
\newblock \href {https://doi.org/10.18653/v1/2020.coling-main.297} {An unsupervised method for learning representations of multi-word expressions for semantic classification}.
\newblock In \emph{Proceedings of the 28th International Conference on Computational Linguistics}, pages 3346--3356, Barcelona, Spain (Online). International Committee on Computational Linguistics.

\bibitem[{Wada et~al.(2023)Wada, Matsumoto, Baldwin, and Lau}]{wada-etal-2023-unsupervised}
Takashi Wada, Yuji Matsumoto, Timothy Baldwin, and Jey~Han Lau. 2023.
\newblock \href {https://doi.org/10.18653/v1/2023.findings-acl.290} {Unsupervised paraphrasing of multiword expressions}.
\newblock In \emph{Findings of the Association for Computational Linguistics: ACL 2023}, pages 4732--4746, Toronto, Canada. Association for Computational Linguistics.

\bibitem[{Wolf et~al.(2020)Wolf, Debut, Sanh, Chaumond, Delangue, Moi, Cistac, Rault, Louf, Funtowicz, Davison, Shleifer, von Platen, Ma, Jernite, Plu, Xu, Le~Scao, Gugger, Drame, Lhoest, and Rush}]{wolf-etal-2020-transformers}
Thomas Wolf, Lysandre Debut, Victor Sanh, Julien Chaumond, Clement Delangue, Anthony Moi, Pierric Cistac, Tim Rault, Remi Louf, Morgan Funtowicz, Joe Davison, Sam Shleifer, Patrick von Platen, Clara Ma, Yacine Jernite, Julien Plu, Canwen Xu, Teven Le~Scao, Sylvain Gugger, and 3 others. 2020.
\newblock \href {https://doi.org/10.18653/v1/2020.emnlp-demos.6} {Transformers: State-of-the-art natural language processing}.
\newblock In \emph{Proceedings of the 2020 Conference on Empirical Methods in Natural Language Processing: System Demonstrations}, pages 38--45, Online. Association for Computational Linguistics.

\bibitem[{Wray(2002)}]{wray2002formulaic}
Alison Wray. 2002.
\newblock \emph{Formulaic language and the lexicon}, volume~20.
\newblock Cambridge University Press Cambridge.

\bibitem[{Yang et~al.(2025)Yang, Li, Yang, Zhang, Hui, Zheng, Yu, Gao, Huang, Lv et~al.}]{qwen3technicalreport}
An~Yang, Anfeng Li, Baosong Yang, Beichen Zhang, Binyuan Hui, Bo~Zheng, Bowen Yu, Chang Gao, Chengen Huang, Chenxu Lv, and 1 others. 2025.
\newblock Qwen3 technical report.
\newblock \emph{arXiv preprint arXiv:2505.09388}.

\bibitem[{Yang et~al.(2026)Yang, Liu, Li, Bai, Chen, Chen, Duan, Dong, Hu, Jia et~al.}]{yang2026onemillion}
Qianyu Yang, Yang Liu, Jiaqi Li, Jun Bai, Hao Chen, Kaiyuan Chen, Tiliang Duan, Jiayun Dong, Xiaobo Hu, Zixia Jia, and 1 others. 2026.
\newblock $\$$onemillion-bench: How far are language agents from human experts?
\newblock \emph{arXiv preprint arXiv:2603.07980}.

\bibitem[{Yu et~al.(2024)Yu, Zhang, Tiwari, and Wang}]{10.1145/3664194}
Fei Yu, Hongbo Zhang, Prayag Tiwari, and Benyou Wang. 2024.
\newblock \href {https://doi.org/10.1145/3664194} {Natural language reasoning, a survey}.
\newblock \emph{ACM Comput. Surv.}, 56(12).

\bibitem[{Zeng and Bhat(2022)}]{zeng-bhat-2022-getting}
Ziheng Zeng and Suma Bhat. 2022.
\newblock \href {https://doi.org/10.1162/tacl_a_00510} {Getting {BART} to ride the idiomatic train: Learning to represent idiomatic expressions}.
\newblock \emph{Transactions of the Association for Computational Linguistics}, 10:1120--1137.

\bibitem[{Zeng et~al.(2023)Zeng, Cheng, Nanniyur, Zhou, and Bhat}]{zeng-etal-2023-iekg}
Ziheng Zeng, Kellen Cheng, Srihari Nanniyur, Jianing Zhou, and Suma Bhat. 2023.
\newblock \href {https://doi.org/10.18653/v1/2023.emnlp-main.881} {{IEKG}: A commonsense knowledge graph for idiomatic expressions}.
\newblock In \emph{Proceedings of the 2023 Conference on Empirical Methods in Natural Language Processing}, pages 14243--14264, Singapore. Association for Computational Linguistics.

\bibitem[{Zhang et~al.(2020)Zhang, Kishore, Wu, Weinberger, and Artzi}]{zhang2019bertscore}
Tianyi Zhang, Varsha Kishore, Felix Wu, Kilian~Q. Weinberger, and Yoav Artzi. 2020.
\newblock \href {https://openreview.net/forum?id=SkeHuCVFDr} {Bertscore: Evaluating text generation with bert}.
\newblock In \emph{International Conference on Learning Representations}.

\bibitem[{Zhou et~al.(2021)Zhou, Gong, and Bhat}]{zhou-etal-2021-pie}
Jianing Zhou, Hongyu Gong, and Suma Bhat. 2021.
\newblock \href {https://doi.org/10.18653/v1/2021.mwe-1.5} {{PIE}: A parallel idiomatic expression corpus for idiomatic sentence generation and paraphrasing}.
\newblock In \emph{Proceedings of the 17th Workshop on Multiword Expressions (MWE 2021)}, pages 33--48, Online. Association for Computational Linguistics.

\bibitem[{Zhou et~al.(2022)Zhou, Zeng, Gong, and Bhat}]{zhou2022idiomatic}
Jianing Zhou, Ziheng Zeng, Hongyu Gong, and Suma Bhat. 2022.
\newblock Idiomatic expression paraphrasing without strong supervision.
\newblock In \emph{Proceedings of the AAAI Conference on Artificial Intelligence}, volume~36, pages 11774--11782.

\end{thebibliography}

\clearpage
\appendix
\section*{Appendix}
\hypersetup{
  linkcolor=black
}
\startcontents[sections]
\printcontents[sections]{l}{1}{\setcounter{tocdepth}{2}}

\newpage

\section{Semantic Gloss for Lexical Functions}
\label{sec:appendix-lf_semantic_gloss}
In recent years, there has been an increasing interest in assigning lexical functions as labels to annotated \textsc{MwE} in the sense of the meaning-text theory \citep{mel2023general}. The lexical function is a multi-valued function, which $\mathit{f}$ associates a lexical unit $\mathit{L}$ with a set $\mathit{f(L)}$ of lexical expressions.

As seen in Table \ref{tab:lexfunc-with-semantic-gloss}, we constructed a collection of the representative lexical functions with their semantic glosses from the existing work. We compiled the prompts with the task descriptions.

\section{Additional Details of Datasets}\label{sec:appendix-additional-experiment-details}

\subsection{Idiomacity Detection}\label{sec:appendix-idiom-detection}
In the initial dataset\footnote{\url{https://github.com/H-TayyarMadabushi/AStitchInLanguageModels}} proposed by \citep{tayyar-madabushi-etal-2021-astitchinlanguagemodels-dataset}, there exists three or four possible meanings (\textit{i.e.}, interpretation) for each instance. For instances with only three interpretations, we add the option ``None of the above'' to keep consistency to the four-choices form. We deduplicate according to the unique (idiom, choice) pair for all instances. As a result, we collate 273 examples (\textit{cf.} Table \ref{tab:task-list}). Figure \ref{fig:data-example-idiomacity-detection} shows an example of data.
\begin{figure}[ht]
    \centering
    \fbox{\parbox{0.92\linewidth}{
        [Context] There is also a covered pavilion. It is located next to \textit{\dashuline{Silver Lining}} Tire Recycling. The hours are 6:00 am to 10:00 pm, year round.
        \\ \\
        \text{[Choices]}\\
        (A) grey lining\hfill\redcross \\(B) unexpected advantage\hfill\redcross \\ \textbf{(C) Proper Noun}\hfill\greencheck \\ (D) Meta Usage\hfill\redcross
    }}
    \caption{A data example of idiomacity detection (IED).}
    \label{fig:data-example-idiomacity-detection}
\end{figure}

\subsection{Idiom Extraction}\label{sec:appendix-idiom-extraction}
The original dataset\footnote{\url{https://github.com/Babelscape/ID10M}} consists of instances with or without idiom $\mathcal{IE}$. Since the inference-only experiments comprise most of our work, we filter out all the examples without the $\mathcal{IE}$ existing to increase the coverage diversity of idioms; then, we deduplicate according to the unique item of the occurred $\mathcal{IE}$. The final prepared test set consists of 447 examples with a unique item of $\mathcal{IE}$ existing in each. Figure \ref{fig:data-example-idiom-extraction} shows an example of data.
\begin{figure}[ht]
    \centering
    \fbox{\parbox{0.9\linewidth}{
        [Context] In the screenplay by Lorenzo Semple Jr. , and David Rayfiel , Turner very early on stumbles upon the existence of a kind of super - C.I.A. within the C.I.A. , after which his life is \textit{\dashuline{not worth a plug nickel}} .
        \\ \\
        \text{[Idiom]} \textbf{``not worth a plug nickel''}\hfill\faicon{gavel}
    }}
    \caption{A data example of idiom extraction (IEE).}
    \label{fig:data-example-idiom-extraction}
\end{figure}

\subsection{Idiom Interpretation}\label{sec:appendix-idiom-interpretation}
We collected 916 instances in total from the PIE \citep{zhou-etal-2021-pie}\footnote{\url{https://github.com/zhjjn/MWE_PIE}} and \citep{chakrabarty2022s}\footnote{\url{https://github.com/tuhinjubcse/FigurativeNarrativeBenchmark}}, after deduplication by occurred items of idiom $\mathcal{IE}$. Figure \ref{fig:data-example-idiom-interpretation} shows an example of data.
\begin{figure}[ht]
    \centering
    \fbox{\parbox{0.9\linewidth}{
        \text{[Context]} The remission at this stage of having cancer was truly the \textit{\dashuline{turning point}} of her life .\\\\
        \text{[Idiom]} ``turning point'' \\\\
        \text{[Interpretation]} \textbf{``the time of significant change (mostly positive) in situation''}\hfill\faicon{quote-left}
    }}
    \caption{A data example of idiom interpretation (IEI).}
    \label{fig:data-example-idiom-interpretation}
\end{figure}

\subsection{Lexical Collocation Categorization}\label{subsec:lcc-exp}
We collect the collocation data with the annotated labels from the expanded \textsc{LexFunc}\footnote{\url{https://github.com/luisespinosaanke/lexicalcollocations}} \citep{espinosa-anke-etal-2021-evaluating}. We inherited the training and validation sets of the initial data and sampled 50 examples per semantic category from the test set in classification concerning the computation efficiency. Figure \ref{fig:data-example-collocation-categorization} shows an example of data.
\begin{figure}[ht]
    \centering
    \fbox{\parbox{0.9\linewidth}{
        [Context] In genoa, the \textit{\dashuline{violent storm}} knocked down power lines, blacking out the homes of 5,000 residents.
        \\ \\
        \text{[Category]} \textbf{Magn} (strong semantic).\hfill\faicon{link}
    }}
    \caption{A data example is the lexical collocation categorization (LCC) by semantic relations. Note that the taxonomy included in the prompt is omitted here.}
    \label{fig:data-example-collocation-categorization}
\end{figure}

\subsection{Lexical Collocation Extraction}
The initial dataset is collected from \citep{fisas-etal-2020-collfren}\footnote{\url{https://github.com/TalnUPF/CollFrEn}}. We select the English part of the data and perform deduplication to filter out overlap collocations. We downsample 50 instances randomly for each semantic category to form our test set and reuse the training and validation sets of the original data. Figure \ref{fig:data-example-collocation-extraction} shows an example of data. We conduct $\mathcal{LC}$ extraction but not identification task, and not query models to distinguish the base and the collocate to simplify the task in this work.
\begin{figure}[ht]
    \centering
    \fbox{\parbox{0.9\linewidth}{
        [Context] He still gets up the moment the \textit{\dashuline{alarm clock rings}} .
        \\ \\
        \text{[Semantic relation]} Strong or intense degree in the lexical semantic relation. \\\\
        \text{[Collocation]}\\ \textbf{``alarm clock rings''}\hfill\faicon{map-pin}
    }}
    \caption{A data example of lexical collocation extraction (LCE).}
    \label{fig:data-example-collocation-extraction}
\end{figure}

\subsection{Lexical Collocation Interpretation}\label{sec:appendix-collocation-interpretation}
The data\footnote{\label{footnote4}\url{https://github.com/luisespinosaanke/lexicalcollocations}} we used is proposed in \citep{espinosa-anke-etal-2021-evaluating}. We perform random sampling from the original data and get the 400 examples (50 per class) as our test set. We manually annotated and revised the test examples, and get the Cohen's kappa coefficient $\kappa = 0.718$, to confirm the quality. An example of data is shown in Figure \ref{fig:data-example-collocation-interpretation}.

\begin{figure}[ht]
    \centering
    \fbox{\parbox{0.91\linewidth}{
        [Context] Through robert bennett, his lawyer, the president continued friday to call mrs. jones' \textit{\dashuline{baseless accusation}}.
        \\ \\
        \text{[Collocation]} ``baseless accusation'' \\ \\
        \text{[Interpretation]} \textbf{``Groundless claim made without substantiation''}\hfill\faSearch
    }}
    \caption{A data example of lexical collocation interpretation (LCI).}
    \label{fig:data-example-collocation-interpretation}
\end{figure}

\subsection{Noun Compound Compositionality}
The annotated noun compound data is collected from the NCTTI\footnote{\url{https://github.com/marcospln/nctti}} \citep{garcia-etal-2021-assessing}. After data processing, we filtered out the compound without reference context, collated 237 examples, and split them into training, validation, and test sets. Figure \ref{fig:data-example-noun-compound-compositionality} shows an example of data.
\begin{figure}[ht]
    \centering
    \fbox{\parbox{0.92\linewidth}{
        [Context] \textit{\dashuline{Fair play}} incorporates the concepts of friendship, respect for others and always playing in the right spirit.
        \\ \\
        \text{[Noun compound]} ``Fair play''
        \\ \\
        \text{[Choices]}\\
        (A) Compositional\hfill\redcross \\\textbf{(B) Partly compositional}\hfill\greencheck \\ (C) None of the above\hfill\redcross \\ (D) Non-compositional\hfill\redcross
    }}
    \caption{A data example of noun compound compositionality (NCC).}
    \label{fig:data-example-noun-compound-compositionality}
\end{figure}

\subsection{Noun Compound Extraction}
As our beginning, we sampled the test set from the \textsc{ProNCI}\footnote{\url{https://github.com/dair-iitd/pronci}} \citep{kolluru-etal-2022-covid}. We used the training and validation sets to leverage the compositional part of noun compounds in the original dataset. We randomly sampled from the test set to form the new test set with 720 examples. We demonstrate a data example in Figure \ref{fig:data-example-noun-compound-extraction}.
\begin{figure}[ht]
    \centering
    \fbox{\parbox{0.9\linewidth}{
        [Context] The rhombus shape of the patches arose by adaptation to the \textit{\dashuline{Paris fashion}} of the 17th century by Biancolelli.
        \\ \\
        \text{[Noun compound]} \textbf{``Paris fashion''}\hfill\faicon{gavel}
    }}
    \caption{A data example of noun compound extraction (NCE).}
    \label{fig:data-example-noun-compound-extraction}
\end{figure}

\subsection{Noun Compound Interpretation}
We leverage the initial training, validation, and test data splits from \citep{coil-shwartz-2023-chocolate}\footnote{\url{https://github.com/jordancoil/noun-compound-interpretation}}. To provide a context for each noun compound, we use ChatGPT to generate a reference sentence. To verify the quality of synthetic data, we performed a manual inspection, which resulted in  $acc>98\%$. A data example is shown in the figure \ref{fig:data-example-noun-compound-interpretation}.
\begin{figure}[ht]
    \centering
    \fbox{\parbox{0.95\linewidth}{
        \text{[Context]} She used a straightedge to draw a \textit{\dashuline{ruler line}} across the paper, ensuring her graph was perfectly aligned. \\\\
        \text{[Noun compound]} ``ruler line'' \\\\
        \text{[Interpretation]} \textbf{``line drawn with a ruler''}~~~\faicon{quote-left}
    }}
    \caption{A data example of noun compound interpretation (NCI).}
    \label{fig:data-example-noun-compound-interpretation}
\end{figure}

\subsection{\textsc{VMwE} Extraction}
We used the English corpus of PARSEME v1.3\footnote{\url{https://gitlab.com/parseme/parseme_corpus_en}} \citep{savary-etal-2023-parseme}, the existing largest annotated corpora of \textsc{VMwE}. The initial data is used to conduct extraction instead of identification tasks. Figure \ref{fig:data-example-vmwe-extraction} shows an example of the data.
\begin{figure}[ht]
    \centering
    \fbox{\parbox{0.9\linewidth}{
        [Context] Harry tore back across the room as the landing light \textit{\dashuline{clicked on}}.
        \\ \\
        \text{[\textsc{VMwE}]} \textbf{``clicked on''}\hfill\faicon{gavel}
    }}
    \caption{A data example of \textsc{VMwE} Extraction.}
    \label{fig:data-example-vmwe-extraction}
\end{figure}

\section{Example Prompt}\label{sec:example-prompt}

We manually create a unified prompt template for all tasks that can be adapted to each task with specific filling arguments. The prompt format is shown in the Figure \ref{unified-prompt-template}. The detailed prompt for each task can be accessed in our code base\footnote{\url{https://github.com/lexbench/LexBench/tree/main/lexbench/prompts}}. \\

\begin{center}
	\scalebox{0.8}{
		\begin{tcolorbox}[enhanced,attach boxed title to top center={yshift=-3mm,yshifttext=-1mm},
		  colback=white,colframe=black!75!black,colbacktitle=black!80!black,
		  title=Unified Prompt Template,fonttitle=\bfseries\small, boxed title style={size=small,colframe=black!50!black}, fontupper=\footnotesize]
		  \em Assume that you are a linguist who researches \text{\{\{semantic phrases\}\}}. \\ \\
		  You will be given a sentence that contains only an item of \text{\{\{semantic phrase\}\}}. \\ \\
		  Your task is to ... \\ \\
		  Please make sure you read and understand these instructions carefully. \\
		  \begin{DottedTextBox}
		    Few-shot Examples: \\
		    \begin{DottedTextBox}Phrase: \textup{\{\{}an example of the phrase\textup{\}\}} \end{DottedTextBox}
		    Context: \textup{\{\{}a context of the example\textup{\}\}} \\
		    Output: \textup{\{\{}an output of the example\textup{\}\}} \\
		    \textbf{...}
		  \end{DottedTextBox}
		  \begin{DottedTextBox}Phrase: \textup{\{\{}phrase\textup{\}\}}\end{DottedTextBox}
		  Context: \textup{\{\{}context\textup{\}\}} \\ \\
		  Output: \\
		\end{tcolorbox}
	}
\end{center}
\begin{center}
    \begin{minipage}{\linewidth}
        \centering
        \captionof{figure}{Unified prompt template used in the work.}\label{unified-prompt-template}
    \end{minipage}
\end{center}

\section{Oracle Prompt}\label{sec:oracle-prompt-template}
\begin{center}
	\scalebox{0.8}{
		\begin{tcolorbox}[enhanced,attach boxed title to top center={yshift=-3mm,yshifttext=-1mm},
		  colback=white,colframe=black!75!black,colbacktitle=black!80!black,
		  title=Oracle Prompt Template,fonttitle=\bfseries\small, boxed title style={size=small,colframe=black!50!black}, fontupper=\footnotesize]
		  \em Assume that you are a linguist who conducts research on \text{\{\{verbal multiword expressions (VMwEs)\}\}}. \\ \\

		    You will be given a context that includes only one \text{\{\{verbal multiword expression\}\}}. \\ \\
		  Your task is to ... \\ \\
		  Please make sure you read and understand these instructions carefully. \\
		  \begin{DottedTextBox}
		    Few-shot Examples: \\
		    Context: \textup{\{\{}a context of the example\textup{\}\}} \\
		    Output: \textup{\{\{}an output of the example\textup{\}\}} \\
		    \textbf{...}
		  \end{DottedTextBox}
        VMwE Definition:  \textup{\{\{}Definition of VMwE\textup{\}\}} \\
		  \begin{DottedTextBox}
          VMwE Definition Example: \\
          Definition of VMwE: \textup{\{\{}\textit{``Verb-particle construction (VPC) is sometimes called phrasal or phrasal-prepositional verb. The meaning of the VPC is fully or partly non-compositional.''}\textup{\}\}}
          \end{DottedTextBox}
		  Context: \textup{\{\{}context\textup{\}\}} \\ \\
		  Output: \\
		\end{tcolorbox}
	}
\end{center}
\begin{center}
    \begin{minipage}{\linewidth}
        \centering
        \captionof{figure}{Oracle prompt template used in the work.}\label{oracle-prompt-template}
    \end{minipage}
\end{center}


\begin{figure*}[t]
    \centering
    \subfloat{
	    \begin{minipage}{0.313\linewidth}
	    \includegraphics[width=\linewidth]{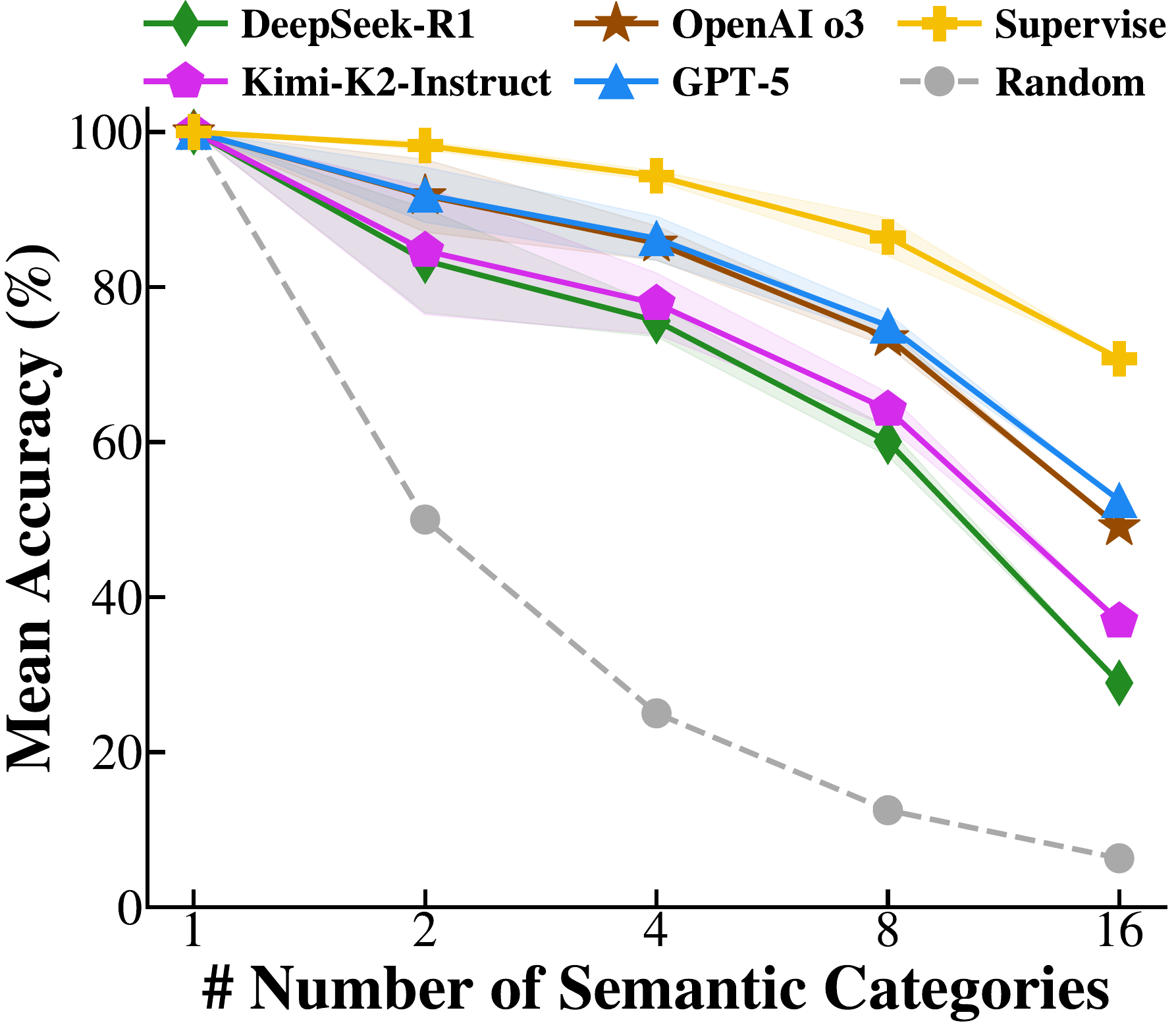}
	    \vspace{-1.1em}
	    \caption{Each model is run with zero-shot prompting in the semantic relation classification with category scaling. Mean accuracy $(\%)$ of different models are average over runs in three sampled sets. For comparative reasons, we also plotted the level of random baseline.}
	    \label{fig:lcc-scaling-comparison}
	    \end{minipage} 
    }
    \hfill\hspace{0em}
    \subfloat{
	    \begin{minipage}{0.313\linewidth}
	    \vspace{.2em}
	    \includegraphics[width=\linewidth]{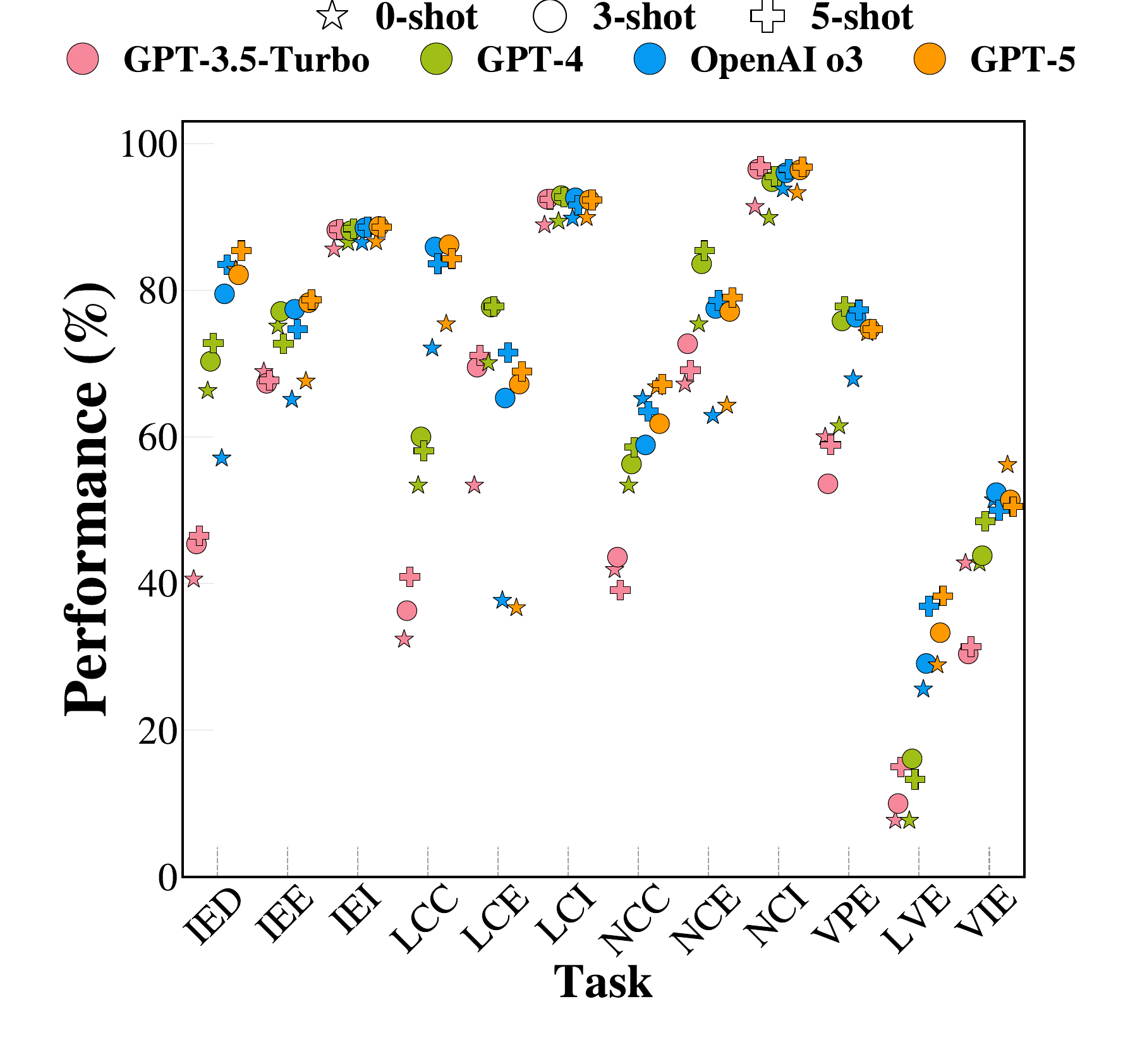}
	    \vspace{-1.1em}
	    \caption{Model performance (GPT-3.5-Turbo, GPT-4, OpenAI o3, GPT-5) across all twelve tasks. Note that the y-axis denotes task-specific metrics, and thus absolute values should not be compared across different tasks.}
	    \label{fig:llama-scaling-comparison}
	    \end{minipage} 
    }
    \hfill\hspace{0em}
    \subfloat{
	    \begin{minipage}{0.313\linewidth}
	    \vspace{-0.7em}
	    \includegraphics[width=\linewidth]{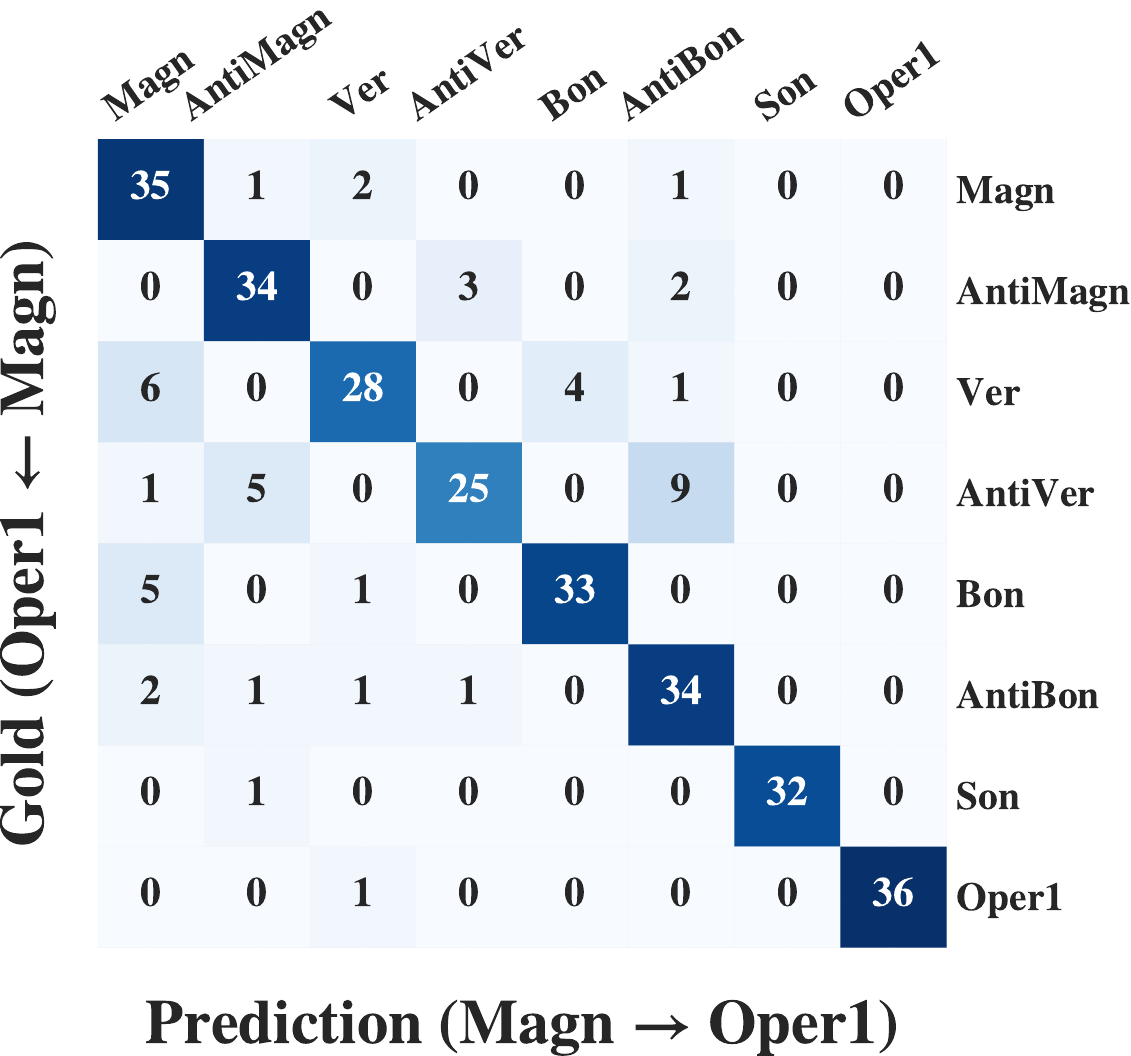}
	    \vspace{-1em}
	    \caption{Confusion matrix for the best-performing model with ICL (GPT-5 in 5-shot setting) in categorizing eight semantic relations described by lexical functions (\textit{cf.} Table \ref{tab:collocation-semantic-relations}). The x-axis denotes the prediction results, and the y-axis represents the gold standards.}
	    \label{fig:lcc-eight-classes-cm}
	    \end{minipage}
    }
\end{figure*}

\section{Additional Experiment Details}\label{sec:additional-exp-details}
\begin{table}[t]
    \centering
    \tiny
    \begin{tabular}{l@{\hspace{0.8em}}c@{\hspace{0.5em}}c@{\hspace{0.8em}}c@{\hspace{0.9em}}c@{\hspace{1em}}c}
        \toprule
            \textbf{Model} & \textbf{\# Params} & \textbf{Arch.} & \textbf{Creator} & \textbf{Public} & \textbf{Post Training} \\
        \midrule
            \hlcelll BERT base$^\dagger$ & 110M & Enc. & Google & \greencheck & FT \\
            \hlcelll BERT large$^\dagger$ & 340M & Enc. & Google & \greencheck & FT \\
            \hlcelll T5 base$^\dagger$ & 220M & Enc.+Dec. & Google &  \greencheck & FT \\
            \hlcelll T5 large$^\dagger$ & 770M & Enc.+Dec. & Google & \greencheck & FT \\
        \midrule
            \hlcellll Qwen3-235B$^\ddagger$ & 235B & Dec.(MoE) & Qwen Team & \greencheck & SFT \\
            \hlcellll DeepSeek-R1$^\ddagger$ & 685B & Dec.(MoE) & DeepSeek-AI & \greencheck & SFT + RL \\
            \hlcellll Kimi-K2-Instruct$^\ddagger$ & 1T & Dec.(MoE) & Kimi Team & \greencheck & SFT \\
            \hlcellll Gemma-3-27B-it$^\ddagger$ & 27B & Dec. & Gemma Team & \greencheck & SFT \\
        \midrule
            \hlcell Gemini-1.0-pro$^\ddagger$ & * & * & Google & \redcross & SFT + RL \\
            \hlcell Gemini-2.5-pro$^\ddagger$ & * & * & Google & \redcross & SFT + RL \\
            \hlcell Claude-Instant-1$^\ddagger$ & * & * & Anthropic & \redcross & SFT + RL \\
            \hlcell Claude-3-Opus$^\ddagger$ & * & * & Anthropic & \redcross & SFT + RL \\
            \hlcell Claude-Sonnet-4.5$^\ddagger$ & * & * & Anthropic & \redcross & SFT + RL \\
            \hlcell GPT-3.5-Turbo$^\ddagger$ & * & * & OpenAI & \redcross & SFT + RL \\
            \hlcell GPT-4$^\ddagger$ & * & * & OpenAI & \redcross & SFT + RL \\
            \hlcell OpenAI o3$^\ddagger$ & * & * & OpenAI & \redcross & SFT + RL \\
            \hlcell GPT-5$^\ddagger$ & * & * & OpenAI & \redcross & SFT + RL \\
        \bottomrule
    \end{tabular}
    \caption{
        A list of LMs tested in this paper:
        ``Public'' indicates whether the model weights are open. In detail, \colorbox{sred}{Light Pink} text delineates the supervised fine-tuned models. \colorbox{sgreen}{Light Green} and \colorbox{sblue}{Light Blue} parts present open-source models and proprietary models, respectively.
        ``Post Training'' indicates whether the model is trained further in some ways after pre-training.
        $^\dagger$We perform trivial full-set fine-tuning for the models.
        $^\ddagger$We use the official API for the model inference.
    }
    \label{tab:models}
    \vspace{-1.4em}
\end{table}
For models accessed via API endpoints, the evaluation probes both zero-shot and few-shot (three- and five-shot) performance. Throughout all experiments, we set the sampling temperature to $\tau = 0$ and employ top‑p decoding \cite{holtzman2019curious} with $p = 1.0$. Inference is accelerated and deployed using vLLM \cite{kwon2023efficient}. For non‑API‑based models, we apply the following configuration. For the sequence classification tasks such as LCC, we employ \texttt{bert-base/large-uncased} as our tuning initiation. Similarly, we construct primary baselines for extraction tasks that leverage the B-I-O scheme to conduct sequence labeling. The training is run with an NVIDIA A100-40GB on Google Colab \citep{bisong2019google}. For interpretation tasks, we use \texttt{t5-base/large} model to conduct vanilla fine-tuning. Additionally, We train all models for a specific number of epochs shown in Table \ref{tab:hyperparams-bert-t5} and perform early stopping over the validation set. Model checkpoints used in our experiment are implemented by PyTorch \citep{paszke2019pytorch}, and Hugging Face Transformers \citep{wolf-etal-2020-transformers}. The input format of the prompt and the few-shot demonstration settings we used during the experiment are shown in Figure \ref{unified-prompt-template}. Since each model has different generation styles, we conduct a pre-run before each test. Then, we develop ad hoc heuristics based on the response generated by models to parse predictions accurately. The perplexity computing in the interpretation tasks is to feed the phrase and its interpretation into the template \textit{``The meaning of phrase \{\{phrase\}\} in context is \{\{interpretation\}\}''}, and then we compute the token-level perplexity by GPT-2-XL \citep{radford2019language}.

\section{Annotation Guideline}\label{sec:annotation-guideline}
We established the following criteria for compiling the dataset of collocation interpretation (\S\ref{subsec:Task-Definition}).
\begin{enumerate}
  \item \textbf{Objective:} Interpret each lexical collocation in five distinct narratives for comprehensive understanding according to the given context.
  \item \textbf{Dataset Overview:} Contains context and collocations paired with  base and collocate.

  \item \textbf{Annotation Format:} Include collocation, five narratives (N1-N5), and rationale.
  
  \item \textbf{Consistency and Accuracy:} Maintain consistent and accurate interpretations across the five narratives in the same semantic meaning.
\end{enumerate}

\begin{table*}
\centering
\scriptsize
\setlength\extrarowheight{1pt}
\begin{tabularx}{\linewidth}{CCC}
\toprule
\textbf{Lexical Function} & \textbf{Semantic Gloss} & \textbf{Complete Description} \\
\midrule
Magn \citep{mel1998collocations} & Intense, strong degree, an intensifier of semantic relation for base lexeme. & Intensify the base lexeme to a high level, strengthening its semantic relation with the associated concept via the collocate lexeme. \\ \hdashline
AntiMagn \citep{mel1998collocations} & Slight and weak degree, a de-intensifier & Weaken meaning intensity, diminishing the semantic relationship between the base lexeme and its associated concept. \\ \hdashline
Ver \citep{gelbukh2012semantic} & Lat. verus, real, genuine & ``As it should be'', ``Meet the intended requirements of \textit{K}''. \\ \hdashline
AntiVer \citep{mel1998collocations} & Non-genuine & Characterize something as non-genuine, not authentic, not in its intended or proper state, and not meeting the required standards or expectations. \\ \hdashline
Bon \citep{espinosa-anke-etal-2021-evaluating} & Positive & Something is good or in a positive situation. \\ \hdashline
AntiBon \citep{espinosa-anke-etal-2021-evaluating} & Negative & Something is bad or in a negative situation. \\ \hdashline
\midrule
IncepPredPlus \citep{fontenelle1997turning} & Start to increase. & Denote initiating a process or action that leads to an increase or enhancement of something. \\ \hdashline
FinFunc0 \citep{kolesnikova2020automatic} & End.existence & The value means ``the \textit{K} of FinFunc0 ceases to be experienced''. \\ \hdashline
Fact0 \citep{mel1998collocations} & Lat. factum, fact. To fulfil the requirement of \textit{K}, and the argument of this function fulfills its own requirement. & Fulfill the base requirement, do something with the base, and do what you are supposed to do with the base. \\
\midrule
CausFunc0 \citep{gelbukh2012semantic} & The agent does something so that the event denoted by the noun occurs & Do something so that \textit{K} begins occurring. \\ \hdashline
Caus1Func0 \citep{espinosa-anke-etal-2021-evaluating} & Cause the existence. 1st argument. & Bring about something's presence or creation, with the first argument indicating the responsible agent or entity. \\ \hdashline
CausFact0 \citep{rodriguez2003domain} & To cause something to function according to its destination. & Denote causing something to function according to its intended purpose or destination. \\ \hdashline
CausPredMinus \citep{fontenelle1997turning} & Cause to decrease. & Describe the act of causing a decrease or reduction in something. \\ \hdashline
CausFunc1 \citep{gelbukh2012semantic} & The non-agentive participant does something such that the event denoted by the noun occurs. & A person/object, different from the agent of \textit{K}, does something so that \textit{K} occurs and has an effect on the agent of \textit{K}. \\ \hdashline
LiquFunc0 \citep{espinosa-anke-etal-2021-evaluating} & Cause termination of the existence & Cause termination of the existence. \\
\midrule
Son \citep{kolesnikova2020automatic} & Lat. \textit{sonare}: sound. & The \textit{K} is usually a noun, and the value means ``emit a characteristic sound''. \\
\midrule
Oper1 \citep{kolesnikova2020automatic} & Lat. \textit{operari}: perform, do, act something. The subject is as the 1st argument. & Represent a light verb linking the event's first participant (subject) with the event's name (direct object). \\ \hdashline
Oper2 \citep{espinosa-anke-etal-2021-evaluating} & Lat. \textit{operari}: perform, do, act something. The subject is as the 2nd argument. & Represent a light verb linking the event's first participant (subject) with the event's name (indirect object). \\ \hdashline
IncepOper1 \citep{gelbukh2012semantic} & Incep is from Lat. \textit{incipere}: begin. Begin to do, perform, experience, carry out \textit{K}. & Signify the start of an action or event, linking the event's subject with its name using a light verb. \\ \hdashline
FinOper1 \citep{kolesnikova2020automatic} & Fin is from Lat. \textit{finire}: cease. & Terminate doing something. \\ \hdashline
Real1 \citep{rodriguez2003domain} & Fulfill a requirement imposed by the noun or performing an action typical for the noun. & To fulfill the requirement of \textit{K}, to act according to \textit{K}. \\
\midrule
Real2 \citep{kolesnikova2020automatic} & Acting as expected. Something be realized as expected  & \textit{K} that is normally expected of the second participant \\ \hdashline
AntiReal2 \citep{kolesnikova2020automatic} & Not acting as expected. Something not be realized as expected. & The \textit{V} is the negation of an internal element of the argument of this function. \\
\bottomrule
\end{tabularx}
\caption{All lexical functions with their semantic gloss in this paper. The column ``semantic gloss'' provides the definition for each LF, and we use a sentence to describe the complete meaning of LF in column ``Complete Description''. \textit{K} denotes the keyword/base word of a LF, and \textit{V} denotes the value/collocate word of a LF.}
\label{tab:lexfunc-with-semantic-gloss}
\end{table*}

\begin{table*}
\centering
\small
\setlength\tabcolsep{4pt}
\setlength{\extrarowheight}{3pt}
\begin{tabular}{lccccccccccc}
\toprule
\multirowcell{2.4}{\textbf{\textsc{VMwE}}} & \phantom & \multicolumn{3}{c}{\textbf{BERT-base}} & \phantom{c} & \multicolumn{3}{c}{\textbf{BERT-large}} & \phantom & \multirowcell{2.4}{\textbf{\# Support}} & \phantom{c} \\
\cmidrule{3-5}\cmidrule{7-9}
&& \textbf{P} & \textbf{R} & \textbf{F1} && \textbf{P} & \textbf{R} & \textbf{F1} \\
\midrule
IAV && 60.7$_{\text{5.6}}$ & 38.0$_{\text{4.3}}$ & 46.5$_{\text{3.3}}$ && 46.5$_{\text{3.6}}$ & 38.9$_{\text{5.6}}$ & 42.3$_{\text{4.8}}$ && 36.0 \\
LVC.cause && 46.4$_{\text{12.2}}$ & 18.4$_{\text{4.0}}$ & 26.2$_{\text{5.6}}$ && 26.4$_{\text{12.4}}$ & 20.7$_{\text{9.1}}$ & 23.2$_{\text{10.5}}$ && 29.0 \\
LVC.full && 52.1$_{\text{4.4}}$ & 61.1$_{\text{2.0}}$ & 56.1$_{\text{2.2}}$ && 55.2$_{\text{2.5}}$ & 56.8$_{\text{8.4}}$ & 55.9$_{\text{5.4}}$ && 172.0 \\
MVC && 95.9$_{\text{4.0}}$ & 80.5$_{\text{2.0}}$ & 87.5$_{\text{2.8}}$ && 100.0$_{\text{0.0}}$ & 80.5$_{\text{2.0}}$ & 89.2$_{\text{1.2}}$ && 29.0 \\
VID && 52.4$_{\text{5.3}}$ & 36.1$_{\text{0.9}}$ & 42.7$_{\text{1.8}}$ && 63.8$_{\text{5.0}}$ & 36.1$_{\text{1.9}}$ & 46.1$_{\text{2.0}}$ && 108.0 \\
VPC.full && 64.3$_{\text{3.3}}$ & 78.4$_{\text{1.6}}$ & 70.6$_{\text{1.5}}$ && 64.4$_{\text{0.1}}$ & 79.4$_{\text{0.5}}$ & 71.1$_{\text{0.2}}$ && 194.0 \\
VPC.semi && 55.9$_{\text{38.7}}$ & 8.9$_{\text{6.9}}$ & 12.9$_{\text{7.3}}$ && 38.8$_{\text{6.4}}$ & 35.6$_{\text{3.9}}$ & 37.1$_{\text{5.0}}$ && 30.0 \\
\midrule
Micro Avg. && 63.2$_{\text{1.9}}$ & 61.2$_{\text{0.6}}$ & 62.3$_{\text{1.3}}$ && 64.2$_{\text{0.5}}$ & 62.4$_{\text{2.7}}$ & 63.3$_{\text{1.6}}$ && 85.4\\
\bottomrule
\end{tabular}
\caption{We report the full results of \textsc{VMwE} extraction reproduced on MTLB-STRUCT. The performance of all categories are defined in the corpora PARSEME 1.3. The corresponding standard deviation is calculated by the results of three runnings with the selected seeds $\{21, 42, 84\}$.} 
    \label{tab:vmwe-full-results}
\end{table*}


\begin{table*}[!ht]
    \centering
    \small

    \begin{tabular}{c}
      \toprule
      \textbf{Computing Infrastructure}\\1\ $\times$\ A100 40GB GPU (Google Colab) \\ 
      \bottomrule \\
    
    \vspace{3mm}\begin{tabular}{cc}
        \toprule
        \textbf{Hyperparameter} & \textbf{Assignment}  \\
        \midrule
        architecture & BERT-\{base, large\} \\
        \midrule
        tokens per sample & $150$ \\
        \midrule
        batch size & $4,800$ \\
        \midrule
        number of workers & $8$ \\
        \midrule
        learning rate & $3e^{-5}$ \\
        \midrule
        number of epochs & $10$ \\
        \midrule
        save interval (epoch) & $1$ \\
        \midrule
        validation interval (epoch) & $1$ \\
        \midrule
        ratio of warmup steps & $3\%$ \\
        \midrule
        learning rate scheduler & Polynomial decay \\
        \midrule
        learning rate optimizer & Adam \\
        \midrule
        Adam beta weights & $(0.9, 0.99)$ \\
        \midrule
        Adam epsilon & $1e^{-6}$ \\
        \midrule
        weight decay & $0$ \\
        \midrule
        random seed & $21$, $42$, $84$ \\
        \bottomrule
    \end{tabular}
    \vspace{3mm}\begin{tabular}{cc}
        \toprule
        \textbf{Hyperparameter} & \textbf{Assignment}  \\
        \midrule
        architecture & T5-\{base, large\} \\
        \midrule
        tokens per sample & $128$ \\
        \midrule
        batch size & $2,048$ \\
        \midrule
        number of workers & $4$ \\
        \midrule
        learning rate & $5e^{-5}$ \\
        \midrule
        number of epochs & $5$ \\
        \midrule
        save interval (epoch) & $1$ \\
        \midrule
        validation interval (epoch) & $1$ \\
        \midrule
        ratio of warmup steps & $3\%$ \\
        \midrule
        learning rate scheduler & Cosine decay \\
        \midrule
        learning rate optimizer & Adam \\
        \midrule
        Adam beta weights & $(0.9, 0.99)$ \\
        \midrule
        Adam epsilon & $1e^{-6}$ \\
        \midrule
        weight decay & $0$ \\
        \midrule
        random seeds & $21$, $42$, $84$ \\
        \bottomrule
    \end{tabular}
    \vspace{-3mm}
    \\ \bottomrule
    \end{tabular}
    
    \caption{Hyperparameters for finetuning BERT-Taggers and T5 Generators.} 
    \label{tab:hyperparams-bert-t5}
\end{table*}

\begin{table*}[!ht]
\centering
\tiny
\setlength\tabcolsep{4pt}
\setlength{\extrarowheight}{2pt}
\begin{tabular}{lcccccccccccccccc}
\toprule
\multirowcell{2.7}{\textbf{{\scriptsize{M}}ODEL}} & \multicolumn{3}{c}{\textbf{{\scriptsize{I}}DIOM}} & \phantom{c} & \multicolumn{3}{c}{\textbf{{\scriptsize{C}}OLLOCATION}} & \phantom{c} & \multicolumn{3}{c}{\textbf{{\scriptsize{N}}OUN {\scriptsize{C}}OMPOUND}} & \phantom{c} & \multicolumn{3}{c}{\textbf{{\scriptsize{VM}}W{\scriptsize{E}}}} & \phantom{c} \\
\cmidrule{2-4}\cmidrule{6-8}\cmidrule{10-12}\cmidrule{14-16}
& \textbf{IED} & \textbf{IEE} & \textbf{IEI} && \textbf{LCC} & \textbf{LCE} & \textbf{LCI} && \textbf{NCC} & \textbf{NCE} & \textbf{NCI} && \textbf{VPE} & \textbf{LVE} & \textbf{VIE} & \\
\midrule
\makecell{\textbf{{\scriptsize{M}}ETRIC~\raisebox{0.23ex}{$(\%)$}}} & $\textsc{Acc}$ & \hspace{0.05cm}$\textsc{Acc}_s$ & B-S && \textsc{Acc} & \hspace{0.05cm}$\textsc{Acc}_s$ & B-S && $\textsc{Acc}$ & \hspace{0.05cm}$\textsc{Acc}_s$ & B-S && \hspace{0.05cm}$\textsc{Acc}_s$ & \hspace{0.05cm}$\textsc{Acc}_s$ & \hspace{0.05cm}$\textsc{Acc}_s$ \\
\midrule
\underline{\textbf{{{\scriptsize{H}}UMAN}}} & 71.0 & \uline{\textbf{87.0}} & 87.6 && 47.0 & 50.0 & 86.8 && \uline{\textbf{71.0}} & 73.0 & 80.3 && \uline{\textbf{85.0}} & \uline{\textbf{55.0}} & \uline{\textbf{78.0}} \\
\midrule
\vspace{0.07em}\underline{\textbf{{\scriptsize{S}}UPERVISED {\scriptsize{M}}ETHODS}} & & & & & & & & & & & & & & & & \\
\hlcelll \textsc{BERT}$_\textsc{B}$: \textit{fine-tuned} & 85.0 & 66.8 & - && 78.8 & 63.1 & - && 53.6 & 68.5 & - && 68.7 & \textcolor{OliveGreen}{\textbf{52.2}} & 36.1 & \\
\hlcelll \textsc{BERT}$_\textsc{L}$: \textit{fine-tuned} & 85.1 & 67.2 & - && 82.6 & 63.8 & - && 51.5 & 69.1 & - && 74.1 & 41.7 & 34.2 & \\
\hlcelll \textsc{T5}$_\textsc{B}$: \textit{fine-tuned} & - & - & 86.8 && - & - & 87.2  && - & - & \textcolor{rred}{\textbf{89.7}} && - & - & - & \\
\hlcelll \textsc{T5}$_\textsc{L}$: \textit{fine-tuned} & - & - & 87.1 && - & - & 87.7  && - & - & 89.8 && - & - & - & \\
\midrule
\vspace{0.07em}\underline{\textbf{{\scriptsize{P}}ROMPT-BASED {\scriptsize{M}}ETHODS}} & & & & & & & & & & & & & \\

\hdashline
\hlcellll Qwen3-235B: \textit{zero-shot} & 64.1 & 53.6 & 86.7 & & 58.0 & \textcolor{rred}{\textbf{25.9}} & 90.3 & & 52.7 & 45.3 & 93.5 & & 56.8 & 19.4 & 39.1 & \\
\hlcellll DeepSeek-R1: \textit{zero-shot} & 71.1 & 69.4 & \textcolor{rred}{\textbf{85.1}} && 66.6 & 31.5 & 90.2 & & 60.2 & 51.3 & 91.3 && 76.8 & 26.7 & 50.5 & \\
\hlcellll \hspace{0.88cm}$\hookrightarrow \text{+}$ \textit{three-shot} & 79.1 & 70.6 & 88.1 & & 76.4 & 55.6 & 91.6 & & 62.7 & 66.3 & 96.3 & & 74.7 & 26.7 & \textcolor{OliveGreen}{\textbf{59.1}} & \\
\hlcellll \hspace{0.88cm}$\hookrightarrow \text{+}$ \textit{five-shot} & 84.3 & 72.3 & 88.0 & & 76.1 & 64.3 & 91.8 & & 60.6 & 70.7 & 96.6 & & 81.6 & 35.8 & 57.1 & \\
\hlcellll Kimi-K2-Instruct: \textit{zero-shot} & 68.5 & 63.1 & 86.7 & & 68.5 & 34.4 & 90.3 & & 60.6 & 45.4 & 95.6 & & 55.8 & 28.9 & 46.7 & \\
\hlcellll \hspace{0.59cm}$\hookrightarrow \text{+}$ \textit{three-shot} & 77.7 & 68.9 & 88.4 & & 79.0 & 67.9 & 92.8 & & 59.3 & 64.4 & 96.7 & & 79.5 & 39.4 & 43.8 & \\
\hlcellll \hspace{0.59cm}$\hookrightarrow \text{+}$ \textit{five-shot} & 81.7 & 69.6 & 88.2 & & 79.7 & 69.2 & 92.3 & & 64.7 & 63.6 & 97.2 & & 81.1 & 43.3 & 46.7 & \\
\hlcellll Gemma-3-27B-it: \textit{zero-shot} & 55.0 & 57.3 & 86.4 & & 58.0 & 38.4 & 89.5 & & 58.3 & 39.9 & 92.1 & & 66.8 & 19.4 & 38.1 & \\
\hlcellll \hspace{0.59cm}$\hookrightarrow \text{+}$ \textit{three-shot} & 69.6 & 62.0 & 88.1 & & 70.1 & 63.7 & 91.1 & & 56.7 & 57.2 & 95.3 & & 74.1 & 28.3 & 45.7 & \\
\hlcellll \hspace{0.59cm}$\hookrightarrow \text{+}$ \textit{five-shot} & 72.1 & 61.6 & 87.9 & & 70.8 & 68.2 & 90.7 & & 56.2 & 59.2 & 95.9 & & 70.5 & 35.0 & 52.4 & \\

\hdashline
\hlcell Gemini-1.0-pro: \textit{zero-shot} & 56.0 & 77.8 & 86.9 & & 48.5 & 51.8 & 89.5 && \textcolor{rred}{\textbf{38.5}} & 59.0 & 91.8 & & 43.8 & \textcolor{rred}{\textbf{6.7}} & 43.8 & \\
\hlcell Gemini-2.5-pro: \textit{zero-shot} & 55.0 & 65.6 & 87.4 & & 71.5 & 52.1 & 89.4 & & 65.6 & 61.2 & 93.7 & & \textcolor{rred}{\textbf{42.6}} & 27.4 & 42.9 & \\
 \hlcell  Claude-Instant-1: \textit{zero-shot} & 51.2 & 72.2 & 85.7 && 40.5 & 42.6 & 89.7 && 43.2 & 50.9 & 91.9 && 59.2 & 11.6 & 39.0 & \\
 \hlcell \hspace{1cm}$\hookrightarrow \text{+}$ \textit{three-shot} & 47.9 & 60.8 & 86.5 && 49.8 & 54.7 & \textcolor{rred}{\textbf{87.0}} & & 47.8 & 59.1 & 94.1 && 48.9 & 18.8 & 35.5 & \\
 \hlcell \hspace{1cm}$\hookrightarrow \text{+}$ \textit{five-shot} & 52.0 & \textcolor{rred}{\textbf{47.4}} & 87.0 & & 50.1 & 57.7 & 87.1 & & 44.9 & 61.8 & 94.5 && 53.1 & 15.0 & 38.4 & \\
 \hlcell Claude-3-Opus: \textit{zero-shot} & 66.3 & 62.8 & 87.1 && 61.3 & 34.7 & 88.5 & & 50.4 & 36.3 & 91.7 && 67.3 & 28.3 & 42.8 & \\
 \hlcell \hspace{0.88cm}$\hookrightarrow \text{+}$ \textit{three-shot} & 75.8 & 64.8 & 88.1 && 69.5 & 56.7 & 92.8 && 56.7 & 33.6 & 93.1 && 74.7 & 37.2 & 47.6 \\
\hlcell  \hspace{0.88cm}$\hookrightarrow \text{+}$ \textit{five-shot} & 72.8 & 67.1 & 88.2 && 69.8 & 60.0 & 92.8 & & 63.9 & \textcolor{rred}{\textbf{30.9}} & 96.0 && 75.7 & 35.5 & 43.2 \\
 \hlcell Claude-Sonnet-4.5: \textit{zero-shot} & 72.5 & 68.5 & 87.2 & & 67.5 & 40.1 & 88.9 & & 51.0 & 45.1 & 94.4 & & 69.8 & 16.1 & 41.9 & \\
\hlcell \hspace{1cm}$\hookrightarrow \text{+}$ \textit{three-shot} & 77.7 & 72.0 & 88.4 & & 77.1 & 70.5 & 91.8 & & 61.4 & 59.3 & 96.8 & & 76.8 & 30.6 & 42.9 & \\
\hlcell \hspace{1cm}$\hookrightarrow \text{+}$ \textit{five-shot} & 78.0 & 72.0 & 88.5 & & 76.1 & 72.7 & 90.9 & & \textcolor{OliveGreen}{\textbf{70.1}} & 62.1 & \textcolor{OliveGreen}{\textbf{97.6}} & & \textcolor{OliveGreen}{\textbf{82.0}} & 37.2 & 47.6 & \\
 \hlcell  GPT-3.5-Turbo: \textit{zero-shot} & \textcolor{rred}{\textbf{40.6}} & 68.9 & 85.6 && \textcolor{rred}{\textbf{32.4}} & 53.4 & 88.9 & & 41.9 & 67.2 & 91.4 && 60.0 & 7.7 & 42.8 & \\
\hlcell \hspace{0.88cm}$\hookrightarrow \text{+}$ \textit{three-shot} & 45.4 & 67.3 & 88.2 & & 36.3 & 69.5 & 92.4 & & 43.6 & 72.7 & 96.5 & & 53.6 & 10.0 & \textcolor{rred}{\textbf{30.4}} & \\
 \hlcell\hspace{0.88cm}$\hookrightarrow \text{+}$ \textit{five-shot} & 46.5 & 67.7 & 88.3 & & 40.9 & 71.1 & 92.4 & & 39.1 & 69.1 & 96.9 && 58.9 & 15.0 & 31.4 & \\
 \hlcell GPT-4: \textit{zero-shot} & 66.3 & 75.1 & 86.5 & & 53.4 & 70.1 & 89.4 & & 53.4 & 75.4 & 89.9 && 61.5 & 7.7 & 42.8 & \\
\hlcell \hspace{0.88cm}$\hookrightarrow \text{+}$ \textit{three-shot} & 70.3 & 77.1 & 88.1 & & 60.0 & 77.7 & \textcolor{OliveGreen}{\textbf{92.9}} & & 56.3 & 83.6 & 94.8 && 75.8 & 16.1 & 43.8 & \\
\hlcell \hspace{0.88cm}$\hookrightarrow \text{+}$ \textit{five-shot} & 72.8 & 72.7 & 88.4 & & 58.1 & \textcolor{OliveGreen}{\textbf{77.8}} & 92.7 & & 58.6 & \textcolor{OliveGreen}{\textbf{85.4}} & 95.5 && 77.8 & 13.3 & 48.5 & \\

\hlcell OpenAI o3 : \textit{zero-shot} & 57.1 & 65.1 & 86.5 && 72.1 & 37.7 & 89.8 & & 65.2 & 62.9 & 93.8 && 67.9 & 25.6 & 51.4 & \\
\hlcell \hspace{0.88cm}$\hookrightarrow \text{+}$ \textit{three-shot} & 79.5 & 77.4 & 88.5 & & 85.9 & 65.3 & 92.6 & & 58.9 & 77.5 & 96.0 & & 76.3 & 29.1 & 52.4& \\
\hlcell \hspace{0.88cm}$\hookrightarrow \text{+}$ \textit{five-shot} & 83.5 & 74.7 & 88.6 & & 83.6 & 71.5 & 91.6 & & 63.5 & 78.6 & 96.5 & & 77.3 & 36.9 & 50.0 & \\
\hlcell GPT-5: \textit{zero-shot} & 82.8 & 67.6 & 86.6 & & 75.4 & 36.7 & 89.9 & & 66.8 & 64.3 & 93.3  & & 74.2 & 28.9 & 56.2 & \\
\hlcell \hspace{0.88cm}$\hookrightarrow \text{+}$ \textit{three-shot} & 82.1 & 78.3 & \textcolor{OliveGreen}{\textbf{88.7}} & & \textcolor{OliveGreen}{\textbf{86.2}} & 67.2 & 92.3 & & 61.8 & 77.1 & 96.4 & & 74.7 & 33.3 & 51.4 & \\
\hlcell \hspace{0.88cm}$\hookrightarrow \text{+}$ \textit{five-shot} & \textcolor{OliveGreen}{\textbf{85.4}} & \textcolor{OliveGreen}{\textbf{78.7}} & 88.6 & & 84.3 & 68.9 & 92.3 & & 67.2 & 79.0 & 96.8 & & 74.7 & 38.3 & 50.5 & \\

\bottomrule
\end{tabular}
\caption{Complete Experimental Results in \ourbenchmark. ``-'' denotes the model that is unavailable or inappropriate for the task. \uline{\textbf{Digits}} highlight cases in which human scores are higher than those of all evaluated models, serving as a coarse reference.
 \colorbox{sred}{Light Pink} text delineates the baselines with supervised fine-tuning. \colorbox{sgreen}{Light Green} and \colorbox{sblue}{Light Blue} parts present open-source models and proprietary models.}
\vspace{-4ex}
\label{tab:major-exp-results—appendix}
\end{table*}

\begin{table*}
\centering
\footnotesize
\setlength\tabcolsep{4pt}
\setlength{\extrarowheight}{2pt}
\begin{tabular}{lccccc}
\toprule
\textbf{System} & \textbf{Acc@1} & \textbf{Acc@2} & \textbf{Acc@4} & \textbf{Acc@8} & \textbf{Acc@16} \\
\midrule
\underline{\textbf{\textit{Baselines}}} & & & & & \\
Random & 100.0 & 50.0 & 25.0 & 12.5 & 6.3 \\
Majority & 100.0 & 50.0 & 25.0 & 12.5 & 6.3 \\
\midrule
\underline{\textbf{\textit{Small language models}}} & & & & & \\
BERT$_\text{B}$ & 100.0$_\text{0.0}$ & 98.9$_\text{1.9}$ & 89.4$_\text{4.9}$ & 79.9$_\text{6.4}$ & 69.9$_\text{0.0}$ \\
BERT$_\text{L}$ & 100.0$_\text{0.0}$ & \textbf{98.9$_\text{1.0}$} & \textbf{95.8$_\text{1.4}$} & \textbf{83.5$_\text{5.2}$} & \textbf{71.8$_\text{0.0}$} \\
\midrule
\underline{\textbf{\textit{Large language models}}} & & & & & \\
DeepSeek-R1 \\
~~~$\hookrightarrow +$ \textit{0-shot} & 100.0$_\text{0.0}$ & 81.7 $_\text{11.9}$ & 68.9 $_\text{4.2}$ & 49.3 $_\text{4.0}$ & 35.4 $_\text{0.0}$ \\
~~~$\hookrightarrow +$ \textit{3-shot} & 100.0$_\text{0.0}$ & 83.8 $_\text{7.5}$ & 73.7 $_\text{2.4}$ & 52.0 $_\text{6.2}$ & 45.9 $_\text{0.0}$ \\
~~~$\hookrightarrow +$ \textit{5-shot} & 100.0$_\text{0.0}$ & 80.6 $_\text{11.8}$ & 68.4 $_\text{7.8}$ & 51.7 $_\text{7.7}$ & 47.3 $_\text{0.0}$ \\
Kimi-K2-Instruct \\
~~~$\hookrightarrow +$ \textit{0-shot} & 100.0$_\text{0.0}$ & 85.6 $_\text{14.6}$ & 71.1 $_\text{8.2}$ & 50.4 $_\text{4.6}$ & 44.6 $_\text{0.0}$ \\
~~~$\hookrightarrow +$ \textit{3-shot} & 100.0$_\text{0.0}$ & 92.8 $_\text{6.7}$ & 78.3 $_\text{3.8}$ & 61.5 $_\text{7.9}$ & 49.6 $_\text{0.0}$ \\
~~~$\hookrightarrow +$ \textit{5-shot} & 100.0$_\text{0.0}$ & 87.2 $_\text{4.2}$ & 77.8 $_\text{6.9}$ & 63.8 $_\text{5.2}$ & 51.7 $_\text{0.0}$ \\
OpenAI o3  \\ 
~~~$\hookrightarrow +$ \textit{0-shot} & 100.0$_\text{0.0}$ & 93.3 $_\text{8.3}$ & 82.8 $_\text{4.6}$ & 65.4  $_\text{2.5}$ & 53.3  $_\text{0.0}$ \\
~~~$\hookrightarrow +$ \textit{3-shot} & 100.0$_\text{0.0}$ & 91.7 $_\text{5.9}$ & 85.3 $_\text{3.1}$ & 72.6 $_\text{3.4}$ & 62.9 $_\text{0.0}$ \\
~~~$\hookrightarrow +$ \textit{5-shot} & 100.0$_\text{0.0}$ & 91.1 $_\text{5.5}$ & 88.6 $_\text{4.5}$ & 74.0 $_\text{1.6}$ & 64.6 $_\text{0.0}$ \\
GPT-5 \\
~~~$\hookrightarrow +$ \textit{0-shot} & 100.0$_\text{0.0}$ & 92.2 $_\text{6.3}$ & 84.2 $_\text{5.9}$ & 67.7 $_\text{3.4}$ & 56.3 $_\text{0.0}$ \\
~~~$\hookrightarrow +$ \textit{3-shot} & 100.0$_\text{0.0}$ & 92.8 $_\text{9.1}$ & 88.6 $_\text{3.8}$ & \underline{74.6 $_\text{2.8}$} & \underline{65.8 $_\text{0.0}$} \\
~~~$\hookrightarrow +$ \textit{5-shot} & 100.0$_\text{0.0}$ & \underline{94.4 $_\text{6.7}$} & \underline{89.2 $_\text{4.1}$} & 73.5 $_\text{3.6}$ & 65.2 $_\text{0.0}$ \\
\bottomrule
\end{tabular}
\caption{Our best experimental results (avg$_\text{std}$). The mean accuracy scores with their standard deviation are computed by averaging the results of three independent runs with different random seeds. Results of baselines are also provided including random choice as well as the majority of class instances over each sub categorization tasks. The \textbf{Bold} and \underline{underlined} texts denote the best and second-best performance in the specific category, respectively.}
\label{tab:lcc-scaling-exp-detail}
\end{table*}

\iffalse

\iftaclpubformat

\onecolumn

\appendix
\section{Author/Affiliation Options as set forth by MIT Press}
\label{sec:authorformatting}

Option 1. Author’s address is underneath each name, centered.

\begin{quote}\centering
  \begin{tabular}{c}
    \textbf{First Author} \\
    First Affiliation \\
    First Address 1 \\
    First Address 2 \\
    \texttt{first.email@example.com}
  \end{tabular}
  \ 
  \begin{tabular}{c}
    \textbf{Second Author} \\
    Second Affiliation \\
    Second Address 1 \\
    Second Address 2 \\
    \texttt{second.email@example.com}
  \end{tabular}

  \begin{tabular}{c}
    \textbf{Third Author} \\
    Third Affiliation \\
    Third Address 1 \\
    Third Address 2 \\
    \texttt{third.email@example.com}
  \end{tabular}
\end{quote}

Option 2. Author’s address is linked with superscript characters to its name,
author names are grouped, centered.

\begin{quote}\centering
    \textbf{First Author$^\diamond$} \quad \textbf{Second Author$^\dagger$} \quad
    \textbf{Third Author$^\ddagger$}
    \\ \ \\
    $^\diamond$First Affiliation \\
    First Address 1 \\
    First Address 2 \\
    \texttt{first.email@example.com}
     \\ \ \\
     $^\dagger$Second Affiliation \\
    Second Address 1 \\
    Second Address 2 \\
    \texttt{second.email@example.com}
     \\ \ \\
    $^\ddagger$Third Affiliation \\
    Third Address 1 \\
    Third Address 2 \\
    \texttt{third.email@example.com}
\end{quote}
  
\fi

\begin{table*}[t]
\centering
\small
\setlength{\tabcolsep}{6pt}

\begin{tabular}{p{3.2cm} p{3.5cm} p{6.5cm}}
\toprule
\textbf{SP Category} & \textbf{SP Subtype} & \textbf{Representative Literature} \\
\midrule

\multirow{9}{*}{\textbf{Idiomatic Expressions}}
& Opaque & \citet{54311d36-f6a7-378f-b3bb-f2b37bdfb9f1} \\
& Semi-transparent & \citet{54311d36-f6a7-378f-b3bb-f2b37bdfb9f1} \\
& Decomposable & \citet{54311d36-f6a7-378f-b3bb-f2b37bdfb9f1} \\
& Pragmatic & \citet{wray2002formulaic} \\
& Formulaic & \citet{wray2002formulaic} \\
& Figurative & \citet{10.1007/3-540-45715-1_1} \\
& Eventive & \citet{chen-etal-2017-leveraging} \\
& Stative & \citet{Spathas2021} \\
& Property & \citet{fazly2009unsupervised} \\

\midrule

\multirow{8}{*}{\textbf{Lexical Collocations}}
& Intensification & \citet{mel1998collocations} \\
& Veracity & \citet{mel1998collocations} \\
& Evaluation & \citet{mel1998collocations} \\
& Operative & \citet{mel1998collocations} \\
& Causative & \citet{mel1998collocations} \\
& Emission & \citet{mel1998collocations} \\
& Typicality & \citet{mel1998collocations} \\
& Structural pattern & \citet{mel1998collocations} \\

\midrule

\multirow{10}{*}{\textbf{Noun Compounds}}
& Material & \citet{tratz-hovy-2010-taxonomy} \\
& Purpose & \citet{tratz-hovy-2010-taxonomy} \\
& Source & \citet{tratz-hovy-2010-taxonomy} \\
& Topic & \citet{tratz-hovy-2010-taxonomy} \\
& Location & \citet{tratz-hovy-2010-taxonomy} \\
& Temporal & \citet{tratz-hovy-2010-taxonomy} \\
& Agentive & \citet{tratz-hovy-2010-taxonomy} \\
& Part-Whole & \citet{tratz-hovy-2010-taxonomy} \\
& Possessive & \citet{tratz-hovy-2010-taxonomy} \\
& Proper compound & \citet{tratz-hovy-2010-taxonomy} \\

\midrule

\multirow{3}{*}{\textbf{Verbal MWEs}}
& Light Verb & \citet{savary-etal-2017-parseme} \\
& Verb-Particle & \citet{savary-etal-2017-parseme} \\
& Verbal Idiom & \citet{savary-etal-2017-parseme} \\

\bottomrule
\end{tabular}

\caption{Correspondence between the proposed semantic phrase subtypes and related categories in established MWE literature with representative references.}
\label{tab:sp_mwe_mapping}

\end{table*}

\end{document}